\documentclass[a4paper,fleqn]{cas-dc}

\usepackage[numbers]{natbib}
\usepackage{epstopdf}
\usepackage{color}
\usepackage{booktabs} % To thicken table lines
\usepackage{hyperref}
\usepackage{verbatim}
\usepackage{microtype}

\def\tsc#1{\csdef{#1}{\textsc{\lowercase{#1}}\xspace}}
\tsc{WGM}
\tsc{QE}
\tsc{EP}
\tsc{PMS}
\tsc{BEC}
\tsc{DE}

\begin{document}
\let\WriteBookmarks\relax
\def\floatpagepagefraction{1}
\def\textpagefraction{.001}
\shorttitle{}
\shortauthors{Z. Feng et~al.} %% 缩略作者 自己名字， 比如： 张三 = S. Zhang

%% 标题
\title [mode = title]{Cluster-CAM: Cluster-Weighted Visual Interpretation of CNNs' Decision in Image Classification}                      
%%\tnotemark[1,2]

%\tnotetext[1]{This document is the results of the research project funded by  Science and technology project of Xianyang city (2021ZDZX-GY-0001), the National Natural Science Foundation of China (No. 61871301), the National Natural Science Foundation of China (No. 62071349).}

%%\tnotetext[2]{The second title footnote which is a longer text matter to fill through the whole text width and overflow into another line in the footnotes area of the first page.}

%% 作者顺序
%% 1
\author[1]{\textcolor[RGB]{0,0,1}{Zhenpeng Feng}}
\address[1]{School of Electronic Engineering, Xidian University, Xi'an, China}  

%% 2
%\author[1]{\textcolor[RGB]{0,0,1}{Xiyang Cui}}
%\cormark[1]%%通讯作者星标

%%3
\author[1]{\textcolor[RGB]{0,0,1}{Hongbing Ji}}
\cormark[1]%%通讯作者星标
\ead{hbji@xidian.edu.cn}
%\address[1]{School of Electronic Engineering, Xidian University, Xi'an, China}  

\author[2]{\textcolor[RGB]{0,0,1}{Milo\v{s} Dakovi\'c}}
%\fnmark[5]
%\address[2]{Faculty of Electrical Engineering, University of Montenegro, Podgorica, Montenegro}

\author[1]{\textcolor[RGB]{0,0,1}{Xiyang Cui}}
%% 4
\author[1]{\textcolor[RGB]{0,0,1}{Mingzhe Zhu}}
%\fnmark[4]
%\address[1]{School of Electronic Engineering, Xidian University, Xi'an, China}

\author[2]{\textcolor[RGB]{0,0,1}{Ljubi\v{s}a~Stankovi\'c}}
%\fnmark[5]
\address[2]{Faculty of Electrical Engineering, University of Montenegro, Podgorica, Montenegro}

\cortext[cor1]{Corresponding author: Hongbing Ji} %% 首页左下角通讯作者
%%\cortext[cor2]{Principal corresponding author} 

%%\fntext[fn1]{This is the first author footnote. but is common to thirdauthor as well.}
%%\fntext[fn2]{Another author footnote, this is a very long footnote and it should be a really long footnote. But this footnote is not yet sufficiently long enough to make two lines of footnote text.}

%%\nonumnote{This note has no numbers. In this work we demonstrate $a_b$ the formation Y\_1 of a new type of polariton on the interface between a cuprous oxide slab and a polystyrene micro-sphere placed on the slab.}

%%摘要
\begin{abstract}
Despite the tremendous success of convolutional neural networks (CNNs) in computer vision,  the mechanism of CNNs still lacks clear interpretation. Currently, class activation mapping (CAM), a famous visualization technique to interpret CNN's decision, has drawn increasing attention. Gradient-based CAMs are efficient while the performance is heavily affected by gradient vanishing and exploding. In contrast, gradient-free CAMs can avoid computing gradients to produce more understandable results. However, existing gradient-free CAMs are quite time-consuming because hundreds of forward interference per image are required.  
In this paper, we proposed Cluster-CAM, an effective and efficient gradient-free CNN interpretation algorithm. Cluster-CAM can significantly reduce the times of forward propagation by splitting the feature maps into clusters in an unsupervised manner. Furthermore, we propose an artful strategy to forge a cognition-base map and cognition-scissors from clustered feature maps. The final salience heatmap will be computed by merging the above cognition maps.
Qualitative results conspicuously show that Cluster-CAM can produce heatmaps where the highlighted regions match the human's cognition more precisely than existing CAMs. The quantitative evaluation further demonstrates the superiority of Cluster-CAM in both effectiveness and efficiency.	
\end{abstract}

%%%\includegraphics{figs/grabs.pdf} %%图片摘要地址路径

%%高亮
%\item highlights 1.
%\item highlights 2.
%\item highlights 3.

%% 关键词
\begin{keywords}
keywords-1 explainable artificial intelligence\sep
keywords-2 class activation mapping  \sep 
keywords-3 clustering algorithm \sep 
keywords-4 image classification
\end{keywords}

% 此指令为生成标题格式，不可删除
\maketitle  

%% 1.引言
\section{Introduction}\label{sec:1}
Convolutional neural networks (CNNs) have provided a basis for numerous remarkable achievements in various computer vision tasks like, for example, image classification \cite{AlexNet, ResNet, srinivas2021bottleneck, LIU2022193}, object detection \cite{Yolo, zhu2017couplenet,cao2019hierarchical, LIU202327}, and semantic segmentation \cite{liu2022convnet, liang2018symbolic, 9388704}. 
Despite CNNs' extraordinary performance, they still lack a clear interpretation of the inner mechanism \cite{CleverHans, ProbeFeature, SALEEM2022165}. This lack of transparency can indeed be a disqualifying factor in some peculiar scenarios where mistakes in interpretation can jeopardize human life and health, like in medical image processing or autonomous vehicles \cite{ren2021interpreting, ZHAO2019185, 9528915, 8889997}. Therefore, it is highly desirable to find a way to understand and explain what exactly CNNs have learned during the training process \cite{9772740, TAN202258, MACPHERSON2021603}. 

Recently, Class Activation Mapping (CAM), a visual interpretation technique, has drawn increasing attention \cite{SUN2022989, TU2021443}. CAM aims at highlighting salience regions of an input image for CNN’s decision using a linearly weighted combination of feature maps. Vanilla CAM directly utilizes the weight of each feature map after global average pooling (GAP) corresponding to the target class, so it is only available for CNNs with GAP \cite{CAM}. To further extend CAM to more complex CNN structures, numerous modified CAMs are proposed and they can be broadly categorized as: 1) gradient-based CAMs, and 2) gradient-free CAMs. Gradient-based CAMs (e.g. Grad-CAM \cite{GradCAM}, Grad-CAM++ \cite{GradCAMplus}, SmoothCAM++ \cite{omeiza2019smooth}, XGrad-CAM \cite{Xgradcam} etc.) define the weights of each feature map using the average of partial gradient of the predicted score with respect to the feature maps. Gradient-based CAMs are usually computed efficiently. However, their weights lack reasonable explanation and are easily impacted by gradient exploding or vanishing. To address this limitation, some gradient-free CAMs are proposed.  They define an intuitive impact of each feature map on the predicted score instead of using the gradient. Examples of gradient-free CAMs are the Ablation CAM \cite{AblationCAM} and the Score-CAM \cite{ScoreCAM}. Gradient-free CAMs can provide a more explainable weight definition than gradient-based CAMs in most cases. Fig.~\ref{fig: comparison} shows the salience heatmaps produced by several aforementioned CAMs.

Although gradient-free CAMs define the weights more reasonably, they are usually very time-consu\-ming since hundreds of forward propagations per image are required. To improve the efficiency of gradient-free CAMs, Q. Zhang et al. proposed Group-CAM where feature maps are split into several groups \cite{zhang2021group}. In this case, only several forward propagations are needed in computing the weights. Nonetheless, the feature maps are split without any regulation in the Group-CAM. Actually, various feature maps have learned different semantic concepts relevant/irrelevant to the object. Therefore, the feature maps should be split into groups/clusters. Those with similar semantics should be assigned to the same group. Z. Feng et al. proposed SC-SM CAM using spectral clustering to accomplish this goal, particularly for synthetic aperture radar (SAR) images \cite{SCSMCAM}. However, no further analysis or modification of weights is mentioned in the SC-SM CAM \cite{SelfMatchingCAM, SCSMCAM}.

In this paper, we propose a Cluster-CAM, an effective and efficient gradient-free CAM, based on unsupervised clustering. In Cluster-CAM, an unsupervised clustering technique, \mbox{K-means}/spectral clustering is adopted to split feature maps into several clusters. Subsequently, we provide an artful strategy to merge those feature maps into a cognition-base map and a cognition-scissors map which will be combined as the final salience heatmap.
The highlights of this paper are as follows:
\begin{itemize}
	\item We propose a Cluster-CAM, as the first attempt to provide a cluster-weighted CAM framework via unsupervised clustering based on optical images.

	\item We provide a novel and artful weight-forming strategy to merge the cognition-base map and cognition-scissors map. These two maps greatly match the human's cognition and intuition, thus the weights are completely reasonable and understandable.

	\item Cluster-CAM is effective and efficient, which outperforms existing gradient-free CAMs in performance in most cases with significantly lower computing costs.
\end{itemize}

The rest of this paper is organized as follows. Section~\ref{sec:2} introduces the basic knowledge of various CAMs. Section~\ref{sec:3}  elaborates on how to generate salience heatmaps by Cluster-CAM. In Section~\ref{sec:4},
various experiments are implemented to demonstrate the validity of Cluster-CAM and further analyze the experimental results from various aspects. Section~\ref{sec:5} concludes this paper.

\section{Related Work}\label{sec:2}

As discussed in Section~\ref{sec:1},  the key issue in interpreting CNN's decision is to explain what the neural network learned to finish a reasonable inference \cite{ProbeFeature, CleverHans}. 
To visualize what CNN focuses on the input image, numerous interpretation algorithms are proposed \cite{zeiler2014visualizing, simonyan2013deep,CAM,NETWORKDISSECTION}, among which Class Activation Mapping draws the most increasing attention due to its simplicity and good performance.

\noindent\textbf{Vanilla CAM:} B. Zhou et al. firstly proposed the vanilla CAM to produce a salience heatmap by a linearly weighted combination of feature maps, $\mathbf{F}_{n}$, with elements $F_n(i,j), \ n=1,2,...,N$, at the target convolutional layer before classification \cite{CAM}
\begin{align}
	&{M}_c(i,j) = \sum_{n}\alpha^{c}_{n}F_n(i,j)\\   
	&S_c = \sum_{n}\omega^{c}_{n}\sum_{i, j}F_n(i, j) = \sum_{n}\alpha^{c}_{n}\sum_{i, j}F_n(i, j),\label{eq:Sc} 
\end{align}
where ${M}_c(i,j) $ is a heat-map and $S_c$ denotes the predicted score for the target class $c$. Thus,
$\alpha^{c}$ are defined by the weights, $\omega^{c}$, of each feature map corresponding to  $c$-th unit in the classification layer. 
Therefore, CAM is only available with CNNs with global-average pooling following the last convolutional layer. To extend CAM to all CNNs,  many modified CAMs are further proposed by manipulating the definition of weights, which are generally categorized as: 1) gradient-based CAMs; 2) gradient-free CAMs.

\subsection{Gradient-based Class Activation Mapping }\label{subsec: gradCAM}
\noindent\textbf{Gradient-based CAM:}  Selvaraju et al. proposed Grad-CAM to visualize any classification CNN architectures by weighting the
feature maps in a certain convolutional layer with the gradients of the predicted score, $\mathbf{s}_c$ with respect to the elements of $\mathbf{F}_{n}$ \cite{GradCAM}, as 
\begin{align}
	&\alpha^{c, Grad}_{n}  =  \sum_{i,j} \frac{\partial \mathbf{s}_{c}}{\partial {F}_{n}(i,j)},
	%	&\mathbf{M}_{grad}^{c}  =  \sum_{d}\alpha_{n}^{c} \mathbf{F}_{n}^{l}
\end{align}
where different from (\ref{eq:Sc}), $\mathbf{s}_c$ is a sparse vector whose elements are zeros except the $c$-th element, which is equal to $S_c$.
However, the highlighted regions generated by Grad-CAM are usually much smaller than the object. To provide a precise highlighted region, some further modified CAMs are proposed, like Grad-CAM++\cite{GradCAMplus}. A. Chattopadhyay et al proposed  Grad-CAM++ which can produce more precise highlighted locality. Grad-CAM++ assumes different elements in the gradient matrix should have different contributions to features maps, thus an extra factor is introduced to realize this assumption using higher order partial gradient, as:
\begin{equation}
	\boldsymbol{\alpha}^{c, Grad++}_{n}  = \frac{\frac{\partial^{2} \mathbf{s}_{c}}{\partial ({F}_{n}(i,j))^{2}}}{2\frac{\partial^{2} \mathbf{s}_{c}}{\partial ({F}_{n}(i,j))^{2}} + \sum_{a,b}{F}_n(a,b) \frac{\partial^{3}\mathbf{s}_c} {({F}_{n}(i,j))^{3}}}   \sum_{i,j} \frac{\partial \mathbf{s}_{c}}{\partial{F}_{n}(i,j)}.
\end{equation}
However, the gradient, $\frac{\partial\mathbf{s}_c}{\partial\mathbf{F}_n}$, is usually heavily noised or somtimes even all-zero. It is probably because 1) CNN is trained to learn a generalized capability to classify a general concept rather than a specific object. 2) some unreasonable phenomena emerged in CNN's training, like gradient vanishing and gradient exploding.  D. Omeiza et al. proposed Smooth Grad-CAM++ to further suppress the noise. The weights of Smooth Grad-CAM++ are defined using the average of the gradients as:
\begin{gather}
	\alpha^{c, SmoothGrad++}_{n} = \\ \notag
	\frac{\frac{1}{m}\sum_{m=1}^{M}D_1^{n}}{2\frac{1}{m}\sum_{m=1}^{M}D_2^{n} + \sum_{a,b}F_n(a,b) \frac{1}{m}\sum_{m=1}^{M}D_3^{n}}   (\frac{1}{m}\sum_{m=1}^{M}D_1^{n})
\end{gather}
where $D_1^{n} = \sum_{i,j}\frac{\partial \mathbf{s}_c}{\partial F_{n}(i,j)}$, $D_2^{n} = \sum_{i,j}\frac{\partial^{2} \mathbf{s}_c}{\partial (F_{n}(i,j))^{2}}$, and $D_3^{n} = \sum_{i,j}\frac{\partial^{3} \mathbf{s}_c}{\partial (F_{n}(i,j))^{3}}$ when the input is added with random noise for $M$ times ($M$ is a constant integer). This smoothing strategy is intuitive but still rough for some complex CNN structures. To further enhance the interpretability of weights, R. Fu et al. proposed XGrad-CAM by introducing two completely explainable axioms to form the weight:
\begin{align}
	&\alpha^{c, XGrad}_{n}  =  \sum_{i,j} \frac{{F}_n(i,j)}{\sum_{i,j}{F}_n(i,j)} \frac{\partial \mathbf{s}_{c}}{\partial {F}_{n}(i,j)}.
\end{align}
Note neither Smooth Grad-CAM++ nor XGrad CAM can guarantee completely avoiding the above unreasonable phenomena in gradient computing.

\begin{figure*}[t]
	\centering
	{\includegraphics{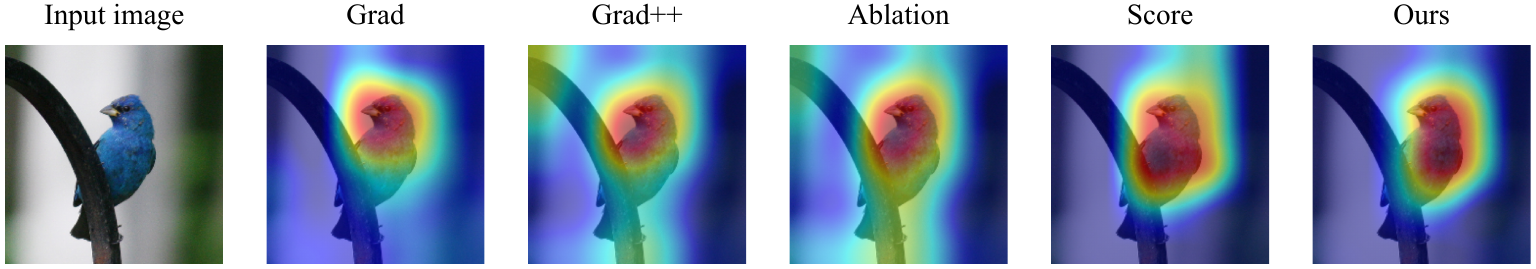}}
	{\includegraphics{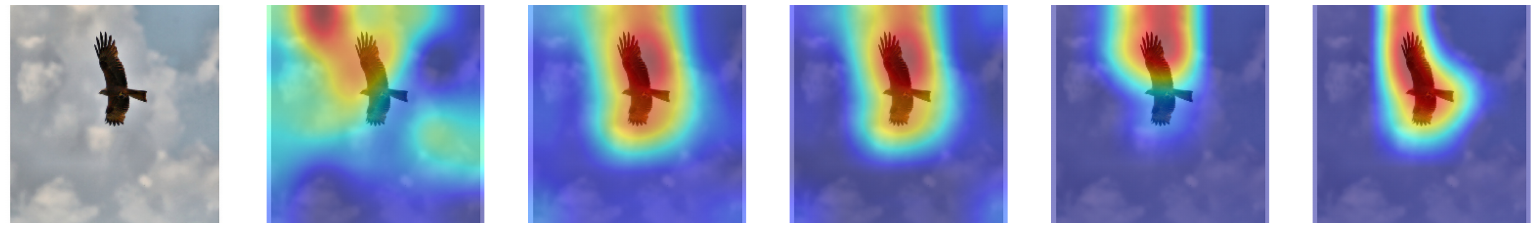}}
	{\includegraphics{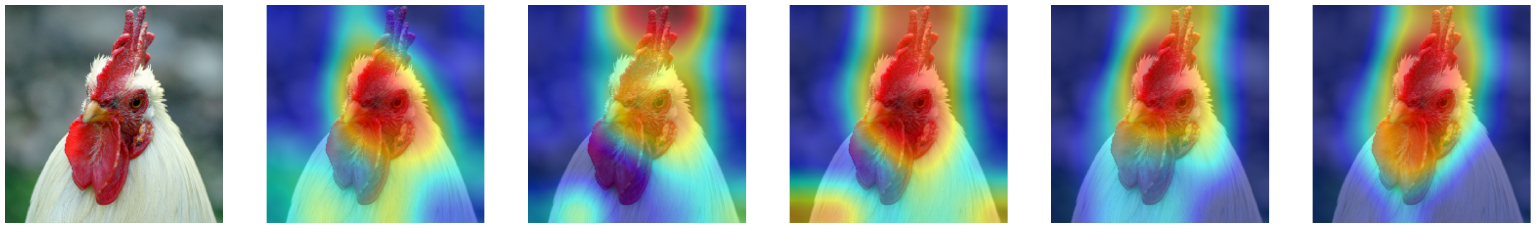}}
	{\includegraphics{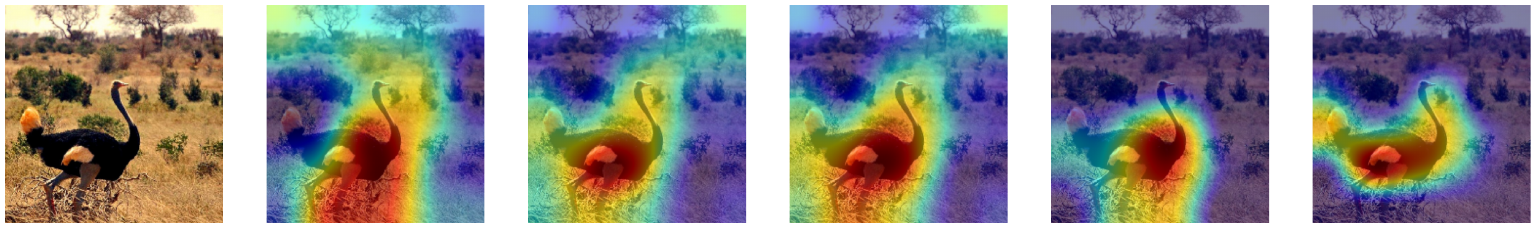}}
	\caption{The heatmaps are produced by different CAMs. The first column is input images (indigo finch, eagle, rooster, and ostrich from top to bottom). The second to fifth columns are heatmaps produced by Grad-CAM, Grad-CAM++, Ablation-CAM, Score-CAM, and Cluster-CAM.}\label{fig: comparison}
\end{figure*}

\subsection{Gradient-free Class Activation Mapping }\label{subsec: gradfreeCAM}
\noindent\textbf{Gradient-free CAM:}To completely solve the problems resulting from gradient computing, some gradient-free CAMs are proposed, i.e., Ablation-CAM \cite{AblationCAM} and Score-CAM \cite{ScoreCAM}, to form the weights using the impact of each feature map on the predicted score instead of using gradient. In Ablation-CAM, the weights are defined as:
\begin{equation}
	\alpha^{c, Ablation}_{n} = \frac{S_{c}-S_{c,n}}{S_{c}}, \label{eq:ablation}
\end{equation}
where $S_{c,n}$ denotes the predicted score for class $c$ when $n$-th feature map is set to zero. In this case, a large weight will be assigned to the current feature map if the removal of it can lead to a dramatic drop in the predicted score ($S_{c}- S_{c,n}$ is a large value) and vice versa. The authors argue that Ablation-CAM is immune to both saturation which marks a filter as important although it is not important, and explosion which marks a filter that has a very small influence as having high importance. Different from Ablation-CAM, Score-CAM considers measuring the impact of the feature map by introducing the input image, $\mathbf{X}$, as
\begin{align}
	\alpha^{c, Score}_{n} &= S_{c}(\mathbf{X} ~ \circ ~ \mathbf{H}_{n}) - S_{c} (\mathbf{X}_{b} ) \label{eq:score}\\
	\mathbf{H}_n &= s(\mathrm{U_p}(\mathbf{F}_{n})),\label{eq:up}
\end{align}
where $\circ$ denotes the element-wise multiplication, $\mathbf{X}_b$ is a baseline image which can be set the input image itself,  $\mathrm{U_p}(\cdot)$ denotes the operation that upsamples $\mathbf{F}_n$
into the input size and $s(\cdot)$ is a normalization function that maps
each element in the input matrix into [$0$, $1$]. $\mathbf{X}~ \circ ~\mathbf{H}_n$ can be deemed as filtering which only passes elements in $\mathbf{X}$ masked by $\mathbf{H}_n$, thus a large weight will be assigned if most target-discriminative are preserved by the current feature map, i.e. $f(\mathbf{X}~ \circ ~\mathbf{H}_n)$ is higher than $S_{c}(\mathbf{X}_b)$ and vice versa. Currently, gradient-free CAMs have drawn more attention than gradient-based CAMs due to their superior performance and explainable definition of weights. However, gradient-free CAMs are much more time-consuming than gradient-based CAMs because hundreds or even thousands of forward interference are required while those gradient-based CAMs only require one forward interference.

\section{Methodology}\label{sec:3}

In this section, we will first introduce some basic concepts on graph-based spectral clustering and K-means.  Then we present the detailed procedures of Cluster-CAM.
%\subsection{K-means}
%{\color{red} You can edit this part about K-means and Spectral Clustering, Prof. Ljubisa.}
\begin{figure*}[t]
	{\includegraphics{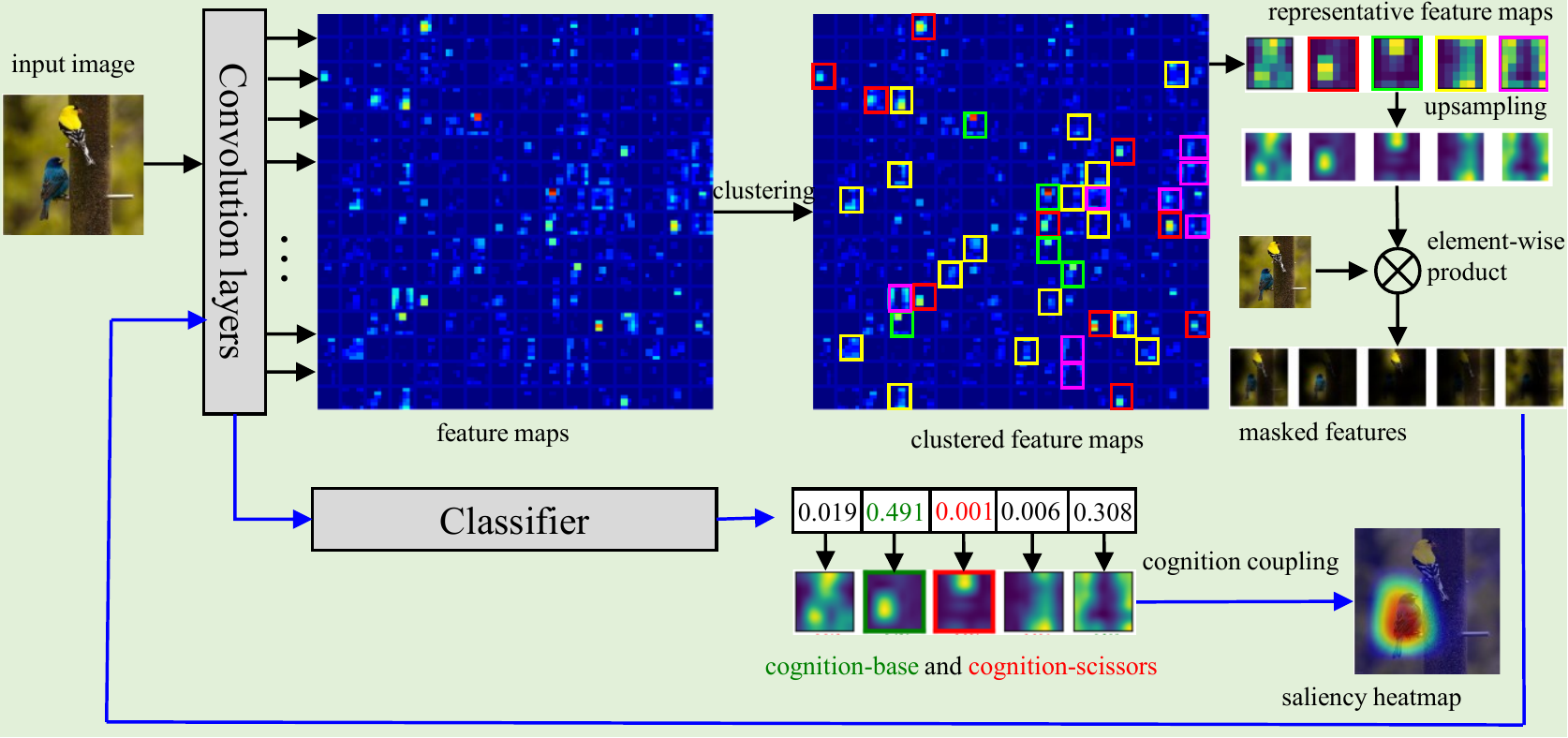}}
	\caption{The flowchart of Cluster-CAM.}\label{fig: flowchart}
\end{figure*}

\subsection{Spectral Clustering and K-means}
Spectral clustering is a widely-used unsupervised clustering algorithm based on graph signal processing  
\cite{spm, SPM2, MA2021108301, SCALZO202383}. Specifically, the processed data (feature maps, $\mathbf{F}_n$) are regarded as vertices in a graph topology. Then the elements, ${S}(i,j)$, of the similarity matrix, $\mathbf{S}$, can be defined:
\begin{align}
	{S}(i,j)= \mathrm{similarity}(\mathbf{F}_i, \mathbf{F}_j) \label{eq:similarity},
\end{align}
where $\mathrm{similarity}(\cdot)$ refers to a function that measures the similarity between two vertices (feature maps). If we use the structural similarity index (SSIM), then it ranges from $0$ (no similarity) to $1$ (identical feature maps). The elements of the weighted adjacency matrix, $\mathbf{A}$, can be defined as:
\begin{equation}
	\left\{
	\begin{array}{lr}
		{A}(i, j) = \exp(-(1-S(i,j)) / \sigma ), ~~\text{if} ~~{S}(i,j)>\theta, &  \\
		{A}(i, j) =  0,  ~~~~~~~~~~~~~~~~~~~~~\text{else,}  &  
	\end{array}
	\right. \label{eq:adjacent}
\end{equation}
where $\theta$ is a threshold to keep the direct edge in the corresponding graph for two neighboring vertices and $\sigma$ is a parameter. Note that,  by definition of similarity, this adjacency matrix is a symmetric matrix resulting in an undirected graph, that is, ${A}(i,j) = {A}(j,i)$. 

The similarity can be defined using the difference between two vertices (feature maps), 	$d(i,j) = ||\mathbf{F}_i-\mathbf{F}_i||$. Then the weighted  adjacency matrix is defined by 
\begin{equation}
	\left\{
	\begin{array}{lr}
		{A}(i,j) = \exp(-d^2(i,j)/ \sigma^2), ~~\text{if} ~~{S}(i,j)>\theta, &  \\
		{A}(i,j) =  0,  ~~~~~~~~~~~~~~~~~~~~~\text{else,}  &  
	\end{array}
	\right. \label{eq:adjacentD}
\end{equation}
where $\theta$ and $\sigma$ have the same role as in (\ref{eq:adjacent}).

In order to produce the vectors for spectral clustering, now we continue and compute the graph Laplacian matrix, $\mathbf{L}$,  as
\begin{align}
	\mathbf{L} = \mathbf{D} - \mathbf{A} \label{eq:laplacian}
\end{align}
where ${D}(i,i) = \sum_{j}{A}(i,j)$ are the elements of the degree matrix $\mathbf{D}$ which is diagonal. 

In practice, the graph Laplacian matrix usually can be normalized, as
\begin{align}
	\mathbf{L}_N = \mathbf{D}^{-\frac{1}{2}} \mathbf{L} \mathbf{D}^{-\frac{1}{2}} = \mathbf{I} - \mathbf{D}^{-\frac{1}{2}} \mathbf{A} \mathbf{D}^{-\frac{1}{2}}. \label{eq:normalizelapalcian}
\end{align}
The clustering results obtained using these two matrices are very similar.

The eigendecomposition of the graph Laplacian 
\begin{align}
	\mathbf{L} = \mathbf{U}^T\boldsymbol{\Lambda}\mathbf{U} \label{eq:eigenlapalcian}.
\end{align}
Results in eigenvectors  $\mathbf{u}_1$, $\mathbf{u}_2$, $\cdots$, $\mathbf{u}_N$ that are the columns of matrix $\mathbf{U}$. The smoothness index of these vectors is equal to the corresponding eigenvalue $\lambda_i$. In clustering the data into two clusters only the eigenvector $\mathbf{u}_2$ is used (Fiedler vector) since the vector  $\mathbf{u}_1$ is omitted as its elements are constant. 

If we want to get a few clusters ($Q$ clusters) then we can use $K$ the smoothest eigenvectors,  $\mathbf{u}_2$, $\mathbf{u}_3$ $\cdots$, $\mathbf{u}_{K+1}$, written in the matrix form as  
\begin{equation}
	\mathbf{B} =  \begin{bmatrix}
		\mathbf{u}_2 & \mathbf{u}_3 &\cdots & \mathbf{u}_{K+1}
	\end{bmatrix}=
	\begin{bmatrix}
		u_{12} & u_{13} &\cdots & u_{1(K+1)} \\
		u_{22} & u_{23} &\cdots  & u_{2(K+1)} \\
		\vdots& \vdots& \ddots& \vdots\\
		u_{N2} & u_{N3} &\cdots & u_{N(K+1)}
	\end{bmatrix}, \label{eq:B}
\end{equation}
where $N$ features with $K$ dimension are considered. 

The clusters are determined based on the $K$-dimensional spectral similarity vectors,  
$\mathbf{q}_1= [ u_{12}, \ u_{13}, \ldots, u_{1(K+1)} ] $, $\mathbf{q}_2= [ u_{22}, \ u_{23}, \ldots, u_{2(K+1)} ] $, $\ldots$, $\mathbf{q}_N= [ u_{N2}, \ u_{N3}, \ldots, u_{N(K+1)} ] $, defined for vertices (features $\mathbf{F}_1$, $\mathbf{F}_2$, $\ldots$ , $\mathbf{F}_N$).

In this way, the dimension of the measuring distance is significantly reduced from the original $N$ dimensional space in $d(i,j) = ||\mathbf{F}_i-\mathbf{F}_i||$ to a very low $K$-dimensional spaces of spectral vectors $\mathbf{q}_n$.

Finally, the clustering result (the data grouped into $Q$ clusters) can be refined using K-means and the Euclidean distance $d(i,j) = ||\mathbf{F}_i-\mathbf{F}_i||$.

Note that the traditional K-means algorithm can be used with an initial random clustering of feature maps into $Q$ clusters, with a slower convergence due to random initialization. In this case, all the feature maps are grouped into $Q$ initial clusters, $\mathbb{Q}_q$, $q=1,2,\dots,Q$. Means of the feature maps are calculated  for each cluster, $M_q= \mathrm{mean}(\mathbf{F}_n, n \in \mathbb{Q}_q)$. The distance of each feature map is checked with respect to each of the mean $M_q$. The feature map is reassigned to the cluster whose mean is the closest to the considered feature map. After all feature maps are considered, the means are recalculated for the new clusters. The procedure is repeated until no feature map changes its cluster.

\subsection{Cluster-CAM}
Now we are ready to introduce spectral clustering and K-means in Cluster-CAM. Here the feature maps, $\mathbf{F}$, represent the vertices in (\ref{eq:similarity}). Take Euclidean distance as similarity measurement, (\ref{eq:similarity}) can be expressed as:
\begin{align}
	{S}(i,j) = \exp{\{-||\mathbf{F}_i-\mathbf{F}_j||}\}, \label{eq:Eucidean}
\end{align}
where a shorter distance means a higher similarity.
By substituting (\ref{eq:Eucidean}) into (\ref{eq:adjacent}), (\ref{eq:laplacian}), (\ref{eq:normalizelapalcian}), and (\ref{eq:B}), we can split $N$ feature maps into $Q$ clusters, $\mathbb{Q}_q$, $q=1,2,\dots,Q$, $Q\ll N$. 
Then we can obtain the $Q$ representative feature maps, $\tilde{\mathbf{F}} = [\mathbf{\tilde{F}}_1, \mathbf{\tilde{F}}_2, \ldots, \mathbf{\tilde{F}}_Q]$,  by calculating the mean of feature maps in each cluster, as
\begin{align}
	\mathbf{\tilde{F}}_{q} = \mathrm{mean} \{ \mathbf{F}_{n}, ~n \in \mathbb{Q}_q \}, \ \ ~ q = 1,\cdots,Q.      
\end{align}

Next we obtain the Hadamard product of $\mathbf{\tilde{F}}$ and $\mathbf{X}$ ($\mathbf{\tilde{F}}$ will be upsampled to the same size of  $\mathbf{X}$). This processing can be deemed as filtering that mainly passes those elements corresponding to large values in  $\mathbf{\tilde{F}}$. The predicted score of each masked image is computed as:
\begin{align}
	\mathbf{y} =& [y_1, y_2, \ldots, y_Q]^{T} \notag \\ 
	 = &  [S_{c, 1} (\mathbf{\tilde{F}}_1 ~ \circ ~ \mathbf{X}), \ldots, S_{c, Q} (\mathbf{\tilde{F}}_Q ~ \circ ~ \mathbf{X})].
\end{align}
In this case, we can obtain the cognition-base map and cognition scissors as
\begin{align}
	\mathbf{\tilde{F}}_{\text{base}} & = \mathbf{F}_{q_{\max}}, \quad  q_{\max}=\mathop{\arg}\limits_{q} \max(\mathbf{y}) \label{eq:base}	\\
	\mathbf{\tilde{F}}_{\text{scissors}} & = \mathbf{F}_{q_{\min}}, \quad q_{\min}=\mathop{\arg}\limits_{q} \min(\mathbf{y}). \label{eq:scissors}
\end{align}
Next, we can semantically couple the cognition-base map and cognition-scissors to form the salience heatmap, as:
\begin{align}
	\mathbf{H}^{Cluster} = \beta \mathbf{\tilde{F}}_{\text{base}} - (1-\beta) \mathbf{\tilde{F}}_{\text{scissors}},
\end{align}
where $\beta\in [0,1]$ is a balance factor to adjust the importance of cognition-base map and cognition-scissors.

\section{Experiments}\label{sec:4}
In this section, we will present and analyze the performance of Cluster-CAM from various perspectives. Firstly we will briefly describe the dataset used in our experiments. Then we verify the superiority of Cluster-CAM to other existing CAMs. 
\subsection{Experimental Setup}
\noindent\textbf{Dataset:} In the following experiments, CNNs are trained on a prevalent benchmark, i.e., ILSVRC \cite{imagenet}. In ILSVRC, there
are around 1.2 million images with $1000$ categories for
training, and $50$ thousand images with $1000$ categories for validation. 

\noindent\textbf{Network Structure:} In this paper, several classic CNNs, AlexNet \cite{AlexNet} and VGG-16 \cite{VGG16}, are used as classification models. Alex-Net is proposed by A. Krizhevsky et al., which consists of $5$ convolutional units (a stack of convolutional layers, ReLU, and max pooling) and  $3$ fully-connected layers. VGG-16 is a very deep CNN with $13$ convolutional layers and $3$ fully-connected layers. VGG-16 has approximately 134M trainable parameters regardless of the output layer.

\begin{figure*}[t]
	\centering
	{\includegraphics{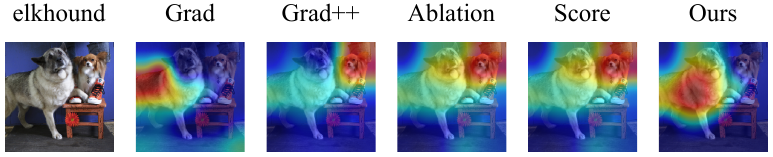}}\hspace{0.2cm}
	{\includegraphics{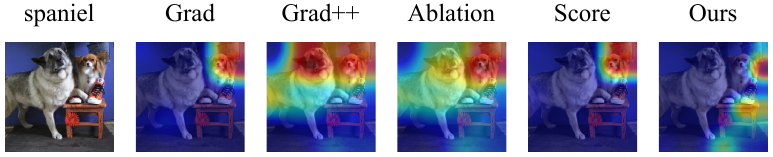}}\vspace{0.25cm}
	{\includegraphics{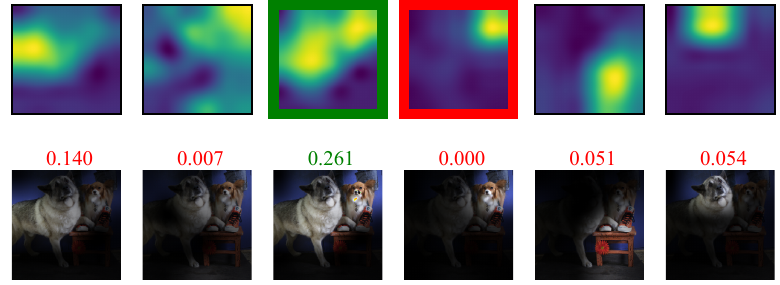}}\hspace{0.13cm}
	{\includegraphics{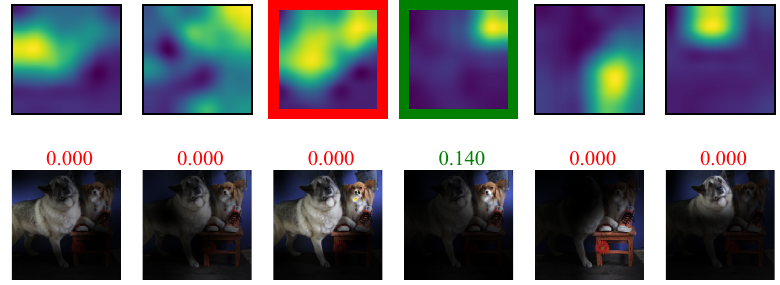}}\vspace{0.4cm}
	{\includegraphics{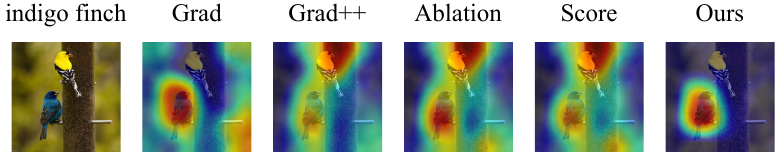}}\hspace{0.2cm}
	{\includegraphics{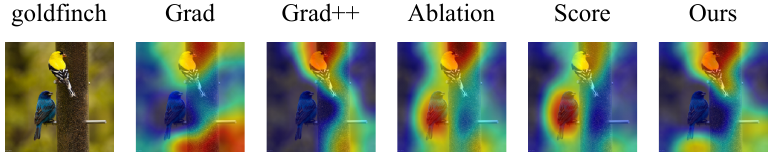}}\vspace{0.25cm}
	{\includegraphics{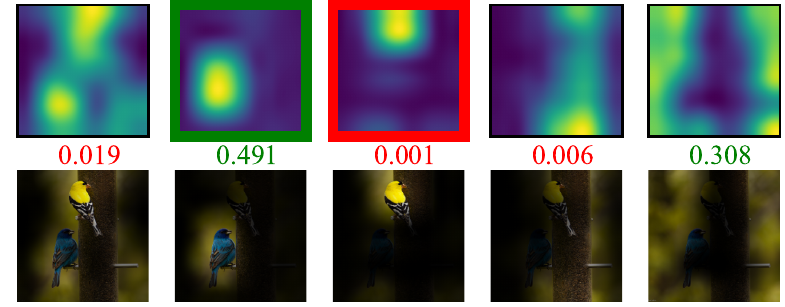}}\hspace{0.13cm}
	{\includegraphics{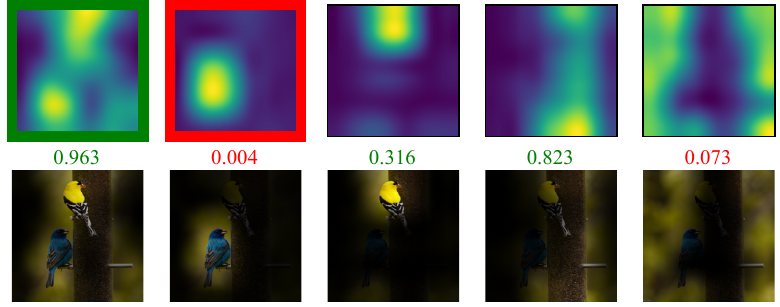}}\vspace{0.4cm}
	\caption{The analysis of feature maps for images of multiple objects. The heatmaps produced by different CAMs for the target label elkhound (first row in the top-left subfigure). The clustered feature maps ( the second row in the top-left subfigure) and corresponding masked images (the third row in the top-left subfigure). Note that the cognition-base map and cognition-scissors are marked by green and red squares, respectively. The predicted score for the current class is provided above each masked image.  The results with the target label spaniel are organized in the same structure (top-right subfigure). The indigo finch and goldfinch are shown in the bottom-left and bottom-right subfigures, respectively.  }\label{fig: multi_objects}\end{figure*}

\subsection{Performance of Discriminative Localization}
Fig.~\ref{fig: comparison} shows the salience heatmaps of different input images (indigo finch, eagle, rooster, and ostrich) by Grad-CAM, Grad-CAM++, Ablation-CAM, Score-CAM, and Cluster CAM. Visually, in comparison to existing CAMs, the highlighted region produced by Cluster-CAM mostly matches human's intuitive understanding of the discriminative part of the specific object. Take the indigo finch as an example, Grad-CAM only highlights the head of the bird, whereas Grad-CAM++ and Ablation-CAM highlight the complete finch but the branch (object-irrelevant information) is also included. Score-CAM and Cluster-CAM highlight the finch body without the branch. But obviously, the region produced by Cluster-CAM matches the profile of the finch more precisely than Score-CAM.

\subsection{CNN's Cognitive Explanation}

\noindent\textbf{Cognition Analysis of Multi-objects Images:} Images of multiple objects are optimal samples to verify the rationality of the cognition-base map and semantic-scissors in (\ref{eq:base}) and (\ref{eq:scissors}). As we discussed in Section~\ref{sec:3}, a reasonable cognition-base map should incorporate the object-relevant information as much as possible, while the corresponding cognition-scissors should include such information as less as better. Therefore, it is necessary to check whether the cognition-base map and cognition-scissors can interchange a multi-objects image if the target class is changed to another object. Fig.~\ref{fig: multi_objects} shows the cognition-base map and cognition-scissors of two muti-objects images as well as the corresponding masked images. There are two types of dogs in the first image, i.e. elkhound (the big gray dog on the left) and spaniel (the tiny brown dog on the right). If the target class is elkhound, the third and the fourth clustered feature map are cognition-base map and cognition-scissors, respectively (marked by green and red squares). It matches human's cognition because cognition-base map incorporates both objects and cognition-scissors only selects the spaniel, thus the highlighted region will only be concentrated on the elkhound, as shown in the first row in the top-left subfigure in Fig.~\ref{fig: multi_objects}. When the target class is changed to the spaniel, the cognition-base map and cognition-scissors are also interchanged, as shown in the third row in the top-right subfigure in Fig.~\ref{fig: multi_objects}. The same phenomenon also emerges in indigo finch and goldfinch, as shown in the bottom subfigures in Fig.~\ref{fig: multi_objects}.
In this case, Cluster-CAM provides solid evidence that CNN's recognition mechanism is similar to human cognition in multiple objects classification.

\begin{figure}[t]
	\centering
	{\includegraphics{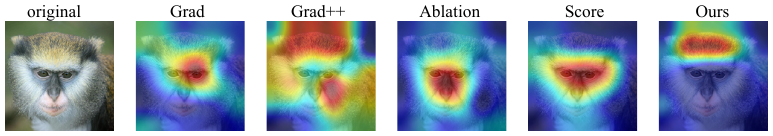}}\vspace{0.3cm}
	{\includegraphics{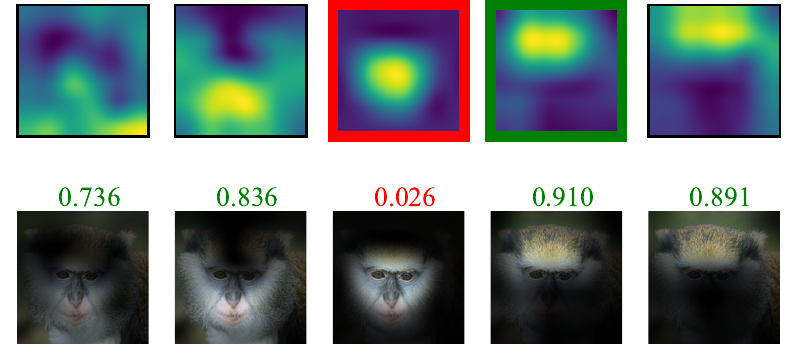}}\vspace{0.5cm}
	{\includegraphics{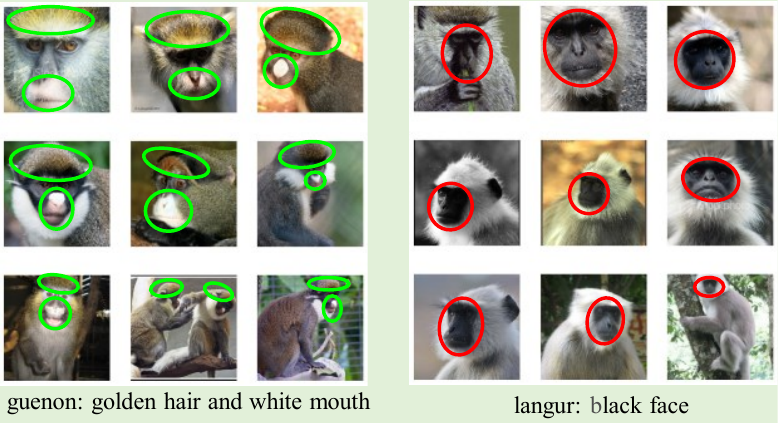}}
	\caption{The cognition-base map (green square) and cognition-scissors (red square) in merged feature maps (top).  The images are masked by corresponding feature masks as well as the predicted score (middle). Nine images of guenon and nine images of langur (bottom).  Note that the discriminative characteristics of guenon (golden hair and white mouth) and langur (black face) are labeled with green and red circles, respectively.}\label{fig: monkey_features}
\end{figure}

\noindent\textbf{Cognition Analysis of Fine-grained Images:}
To further understand how CNN utilizes the learned information to make decisions, we can use CAM to interpret CNNs in fine-grained image classification. 
Fine-grained classification aims to distinguish subordinate categories within entry-level categories. 
Examples include recognizing species of birds such as northern cardinal or  indigo bunting; monkeys such as guenon or langur. 
Fine-grained classification often requires much more detailed information compared with generic object recognition, like the texture of the skin, the thickness of the fur, etc, so CAMs on fine-grained images can tell whether the information is reasonably learned by CNN for classification. 
Fig.~\ref{fig: monkey_features} shows the heatmaps generated by several mentioned CAMs given the input image of a guenon in the first row. 
Interestingly, they focus on completely different parts of the guenon. Grad-CAM and Grad-CAM++ highlight the guenon's eyes and cheek, respectively. 
Ablation-CAM and Score-CAM both highlight the guenon's face, whereas Cluster-CAM only highlights the guenon's forehead. 
Intuitively, Ablation-CAM and Score-CAM seem the most reasonable but the cognition-base map and cognition-scissors clearly show that 
the forehead is the most discriminative part but the face is negative for guenon's classification. It will be understood if we further study the difference in species between guenon and langur. Guenon (widely distributed in Africa) is characterized by blond hair on the forehead and a busty white lip, whereas, langur (distributed in Asia) is characterized by a completely black face. We mark their characteristics by green and red circles in the third row in Fig.~\ref{fig: monkey_features}. It is the reason why the third feature map (face) is deemed as cognition-scissors, i.e., the black face is an interference factor for guenon's categorizing. This example perfectly demonstrates the rationality of Cluster-CAM, particularly the cognition-scissors.

\subsection{Ablation Study}
\begin{figure}[t]
	\centering
	{\includegraphics{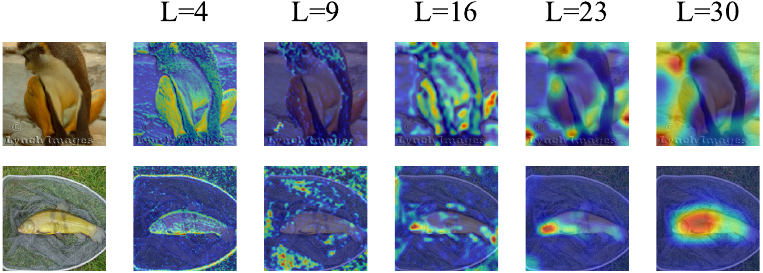}}\vspace{0.12cm}
	{\includegraphics{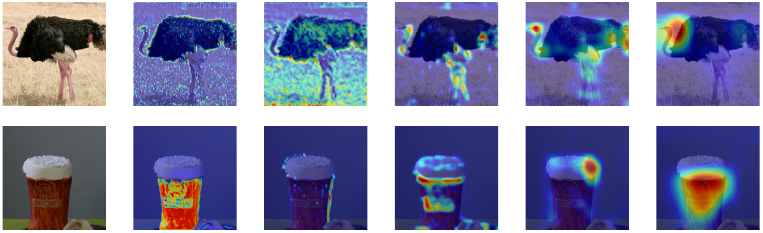}}
	\caption{The heatmaps produced by Cluster-CAM with different numbers of clusters for VGG-16. $L$ refers to the indices of layers in VGG-16.}\label{fig: ablation_cascade}
\end{figure}

\begin{figure}[t]
	\centering
	{\includegraphics{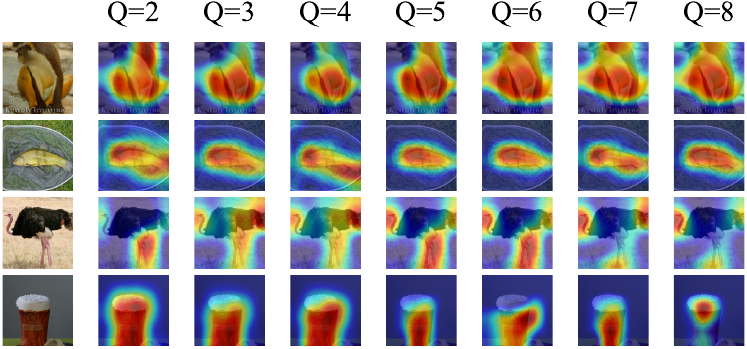}}\vspace{0.2cm}
		{\includegraphics{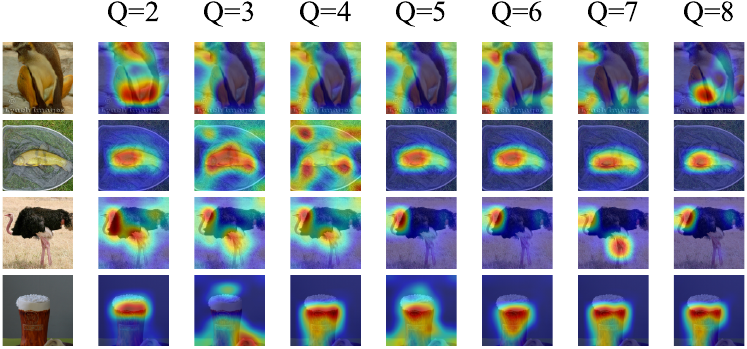}}
	\caption{The heatmaps produced by Cluster-CAM (K-means) with different numbers of clusters for AlexNet (top) and VGG-16(bottom).}\label{fig: ablation_cluster_alex}
\end{figure}

\noindent\textbf{Analysis of Different Layers:} Most CNNs are constructed by a cascade of convolutional blocks (a block consists of convolutional layers, nonlinear activation, pooling operation, etc.). Fig.~\ref{fig: ablation_cascade} shows the salience heatmaps of different convolutional blocks in VGG-16. The results basically match human's intuition that the shallow layers mainly capture some detailed information (e.g., texture and edge), whereas deep layers concentrate on those parts with clearer semantics.

\noindent\textbf{Number of Clusters and CNN Structures:} The number of clusters usually plays a critical role in clustering algorithms. 
Here we vary the cluster number from $2$ to $8$ and present the corresponding salience heatmaps for AlexNet and VGG-16 in Fig.~\ref{fig: ablation_cluster_alex}. It shows the salience heatmaps are sensitive to the number of clusters and are different with CNN models.  
For AlexNet, the highlighted region in heatmaps could be semantic chaos if the feature maps are split into too many or too few clusters. 
It is probably because a large number of clusters may introduce too many detailed patterns of the object and a small number of clusters may directly include background information. Therefore, the number of clusters should be selected as a median value. Note it is only an empirical conclusion and exclusion exists that the optimal value is $2$ for the fourth row (beer) in Fig.\ref{fig: ablation_cluster_alex}. It is probably because the object is simple and in regular shape, thus only two clusters are enough to represent all necessary information. For VGG-16, the highlighted region is more concentrated on a specific part of the object than AlexNet. It is probably because more detailed discriminative information could be captured in VGG-16 which has much deeper layers than AlexNet.

\noindent\textbf{Clustering Method:}
\begin{figure}[t]
	\centering
	{\includegraphics{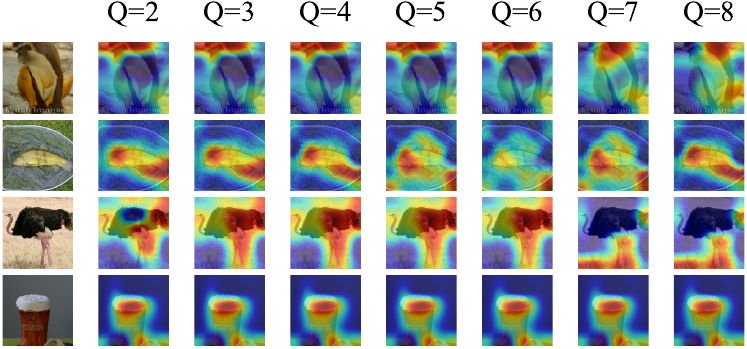}}\vspace{0.2cm}
	{\includegraphics{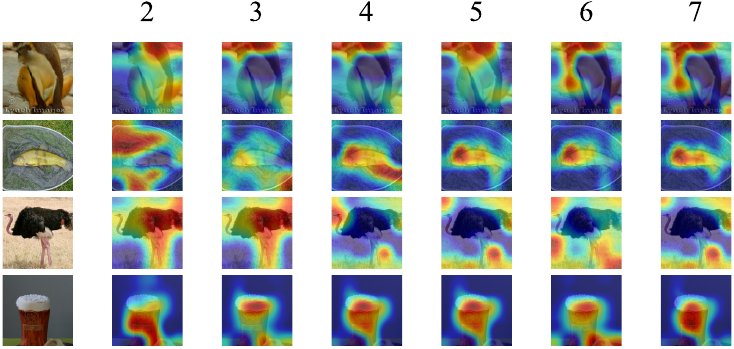}}
	\caption{The heatmaps produced by Cluster-CAM with different numbers of clusters by spectral clustering (top) and heatmaps produced by Cluster-CAM  with different numbers of eigenvectors (bottom).}\label{fig: ablation_cluster_alex_sp}
\end{figure}
In Section~\ref{sec:3}, we introduced two clustering algorithms, i.e., K-means and spectral clustering. Here we take each feature map as a vertex in the graph and use distance to construct the similarity matrix, adjacent matrix, degree matrix, and Laplacian matrix using (\ref{eq:Eucidean}), (\ref{eq:similarity}), (\ref{eq:adjacent}), and (\ref{eq:normalizelapalcian}).  
Fig.~\ref{fig: ablation_cluster_alex_sp} shows the salience heatmaps produced by spectral clustering with different clusters and different eigenvectors. 
It can be observed that the heatmaps are highly related to the number of eigenvectors rather than the clusters.  
Note that the optimal number of eigenvectors is highly related to the image itself, thus careful manipulation of this parameter is required for different objects to obtain the best heatmap. Therefore we will only use K-means to compute the qualitative evaluation metrics in the next section.

\subsection{Qualitative Evaluation}
\subsubsection{Performance Evaluation}
To further evaluate the interpreting performance of Cluster-CAM quantitatively, two widely-used evaluation metrics are adopted in this paper, i.e., confidence drop and increase number \cite{zeiler2014visualizing, zhou2018efficient}.
First of all, let's think about what kind of heatmap can be regarded as a good interpretation of CNN. A natural and intuitive idea is to measure how much the confidence (predicted score) of the target class will drop when the original image is partly occluded according to the heatmap. Specifically, 
for each image, a corresponding explanation map $\mathbf{L}_c$ is generated
by element-wise multiplication of the heatmaps and the current image as in (\ref{eq:up}) and (\ref{eq:score}).

\noindent\textbf{Confidence drop}: This metric compares the average drop of
the model’s confidence for a particular class in an image
after occlusion as:
\begin{equation}
	\mathrm{confidence\_drop} =	\frac{S_{c}(\mathbf{X}) - S_{c}(\mathbf{X} ~ \circ ~ \mathbf{H}_{n})}{S_c(\mathbf{X})},
\end{equation}
For example, assume that CNN predicts an object indigo finch in an image $\mathbf{X}$ with confidence $0.8$. When we input the explanation map, $\mathbf{X}~ \circ ~ \mathbf{H}$, of this image, the CNN’s
confidence in the class indigo finch falls to $0.6$. Then the confidence\_drop would be $25\%$. It means that the most discriminative part ($75\%$) is included in the highlighted region. Confidence drop is expected to be lower for a better CAM and is usually averaged over many images.

\noindent\textbf{Increase number}  measures how many times the CNN’s prediction score for $c$ increased when the masked image is input. Specifically, it happens sometimes that the object is completely included and interference parts are occluded (e.g., the object-irrelevant parts and background) in the highlighted region. In this case, there will be an increase in
the CNN’s predicted score for the class (i.e., confidence drop $< 0$).  This value is computed as a percentage through the whole dataset.

Table~\ref{tab:performancemetric} shows two evaluation metrics of the entire validation set in ILSVRC dataset ($\downarrow$ means the lower value is better and $\uparrow$ means the higher value is better).  These two metrics clearly demonstrate the superiority of Cluster-CAM to other existing CAMs. 
The metrics are computed in Pytorch 1.8.0+cudnn11.1, NVIDIA RTX-3070.

\begin{table}[b]
	\centering
	\caption{Performance Evaluation Metrics. \label{tab:performancemetric}}
	\begin{tabular}{cccc}
		\toprule  % 顶部线
		Method&Confidence drop $\downarrow$&Increase number $\uparrow$\\ 
		\midrule  % 中部线
		Grad-CAM&$17.94$ 	&$19.15$\\	
		Grad-CAM++&$18.44$ 	&$19.75$\\	
		Ablation-CAM&$12.38$ 	&$24.67$\\	
		Score-CAM&$12.21$ 	&$25.48$\\	
		Cluster-CAM&$\mathbf{11.60}$ 	& $\mathbf{26.10}$ \\	
		\bottomrule  % 底部线
	\end{tabular}
\end{table}	

\subsubsection{Efficiency Evaluation}
Here we present two efficiency metrics, i.e., the average computing time and the number of forward propagation (FP) per image in Table~\ref{tab:efficiencycemetric}. It is clear that Cluster-CAM greatly reduces the number of FP compared with Ablation-CAM and Score-CAM. Naturally, a significant improvement in efficiency emerges from Cluster-CAM, i.e., Cluster-CAM is $5.7$ times faster than Ablation-CAM and $12.1$ times faster than Score-CAM. Therefore, Cluster-CAM can obtain better visualization and interpretation performance than gradient-based and gradient-free CAMs with efficiency closer to gradient-based CAMs.

\begin{table}[b]
	\centering
	\caption{Efficiency Evaluation Metrics. Here $Q = 6$ in Cluster-CAM. \label{tab:efficiencycemetric}}
	\begin{tabular}{cccc}
		\toprule  % 顶部线
		Method& Computing time $\downarrow$& number of FP $\downarrow$\\ 
		\midrule  % 中部线

		Grad-CAM&$0.078$ 	&$1$\\	
		Grad-CAM++&$0.141$ 	&$1$\\	
		Ablation-CAM&$2.206$ 	&$256$\\	
		Score-CAM&$4.647$ 	&$256$\\	
		Cluster-CAM&$0.382$ 	&$6$ \\	
		\bottomrule  % 底部线
	\end{tabular}
\end{table}

\section{Conclusion}\label{sec:5}
In this paper, we proposed Cluster-CAM, an effective and efficient CNN interpretation technique based on unsupervised clustering algorithms. Cluster-CAM is the first attempt to comprehensively analyze how to split feature maps into different groups and provide an artful strategy to remove the object-irrelevant elements by defining cognition-scissors. In Cluster-CAM, only several times of forward propagation is required per image while it is usually more than hundreds for other gradient-free CAMs. Qualitative and quantitative experimental results verified Cluster-CAM can obtain even better performance than gradient-free CAMs with much lower computing time.

\section*{Data Availability Statements}\label{res:sec6}
ILSVRC dataset can be downloaded from the website\\
\href{https://www.image-net.org/challenges/LSVRC/}{https://www.image-net.org/challenges/LSVRC/}.

% 致谢
\section*{Acknowledgments}\label{res:sec7}
This work is funded by the National Natural Science Foundation of China (Grant No. 62276204, 61871301, 62071349),
Project 2021ZDZX-GY-0001, science and technology project of Xianyang city.

%The authors are thankful to Prof. Milo\v s Dakovi\'c for the help in the preparation of this manuscript.

\bibliographystyle{cas-model2-names}

\bibliography{References}

\begin{thebibliography}{43}
\expandafter\ifx\csname natexlab\endcsname\relax\def\natexlab#1{#1}\fi
\providecommand{\url}[1]{\texttt{#1}}
\providecommand{\href}[2]{#2}
\providecommand{\path}[1]{#1}
\providecommand{\DOIprefix}{doi:}
\providecommand{\ArXivprefix}{arXiv:}
\providecommand{\URLprefix}{URL: }
\providecommand{\Pubmedprefix}{pmid:}
\providecommand{\doi}[1]{\href{http://dx.doi.org/#1}{\path{#1}}}
\providecommand{\Pubmed}[1]{\href{pmid:#1}{\path{#1}}}
\providecommand{\bibinfo}[2]{#2}
\ifx\xfnm\relax \def\xfnm[#1]{\unskip,\space#1}\fi
%Type = Inproceedings
\bibitem[{Cao et~al.(2019)Cao, Pang, Han and Li}]{cao2019hierarchical}
\bibinfo{author}{Cao, J.}, \bibinfo{author}{Pang, Y.}, \bibinfo{author}{Han,
  J.}, \bibinfo{author}{Li, X.}, \bibinfo{year}{2019}.
\newblock \bibinfo{title}{Hierarchical shot detector}, in:
  \bibinfo{booktitle}{Proceedings of the IEEE/CVF international conference on
  computer vision}, pp. \bibinfo{pages}{9705--9714}.
%Type = Inproceedings
\bibitem[{Chattopadhay et~al.(2018)Chattopadhay, Sarkar, Howlader and
  Balasubramanian}]{GradCAMplus}
\bibinfo{author}{Chattopadhay, A.}, \bibinfo{author}{Sarkar, A.},
  \bibinfo{author}{Howlader, P.}, \bibinfo{author}{Balasubramanian, V.N.},
  \bibinfo{year}{2018}.
\newblock \bibinfo{title}{{Grad-CAM++}: Generalized gradient-based visual
  explanations for deep convolutional networks}, in: \bibinfo{booktitle}{In
  Proceedings of 2018 IEEE Winter Conference on Applications of Computer Vision
  (WACV)}, \bibinfo{organization}{IEEE}. pp. \bibinfo{pages}{839--847}.
%Type = Article
\bibitem[{Chen et~al.(2022)Chen, Jin, Jin, Zhu and Chen}]{9388704}
\bibinfo{author}{Chen, H.}, \bibinfo{author}{Jin, Y.}, \bibinfo{author}{Jin,
  G.}, \bibinfo{author}{Zhu, C.}, \bibinfo{author}{Chen, E.},
  \bibinfo{year}{2022}.
\newblock \bibinfo{title}{Semisupervised semantic segmentation by improving
  prediction confidence}.
\newblock \bibinfo{journal}{IEEE Transactions on Neural Networks and Learning
  Systems} \bibinfo{volume}{33}, \bibinfo{pages}{4991--5003}.
\newblock \DOIprefix\doi{10.1109/TNNLS.2021.3066850}.
%Type = Inproceedings
\bibitem[{Deng et~al.(2009)Deng, Dong, Socher, Li, Li and Fei-Fei}]{imagenet}
\bibinfo{author}{Deng, J.}, \bibinfo{author}{Dong, W.},
  \bibinfo{author}{Socher, R.}, \bibinfo{author}{Li, L.J.},
  \bibinfo{author}{Li, K.}, \bibinfo{author}{Fei-Fei, L.},
  \bibinfo{year}{2009}.
\newblock \bibinfo{title}{Imagenet: A large-scale hierarchical image database},
  in: \bibinfo{booktitle}{In proceedings of 2009 IEEE Conference on Computer
  Vision and Pattern Recognition (CVPR)}, pp. \bibinfo{pages}{248--255}.
\newblock \DOIprefix\doi{10.1109/CVPR.2009.5206848}.
%Type = Article
\bibitem[{Feng et~al.(2021a)Feng, Ji, Stankovi{\'c}, Fan and Zhu}]{SCSMCAM}
\bibinfo{author}{Feng, Z.}, \bibinfo{author}{Ji, H.},
  \bibinfo{author}{Stankovi{\'c}, L.}, \bibinfo{author}{Fan, J.},
  \bibinfo{author}{Zhu, M.}, \bibinfo{year}{2021}a.
\newblock \bibinfo{title}{{SC-SM CAM}: An efficient visual interpretation of
  {CNN} for {SAR} images target recognition}.
\newblock \bibinfo{journal}{Remote Sensing} \bibinfo{volume}{13},
  \bibinfo{pages}{4139}.
%Type = Article
\bibitem[{Feng et~al.(2021b)Feng, Zhu, Stankovi{\'c} and Ji}]{SelfMatchingCAM}
\bibinfo{author}{Feng, Z.}, \bibinfo{author}{Zhu, M.},
  \bibinfo{author}{Stankovi{\'c}, L.}, \bibinfo{author}{Ji, H.},
  \bibinfo{year}{2021}b.
\newblock \bibinfo{title}{Self-matching {CAM}: A novel accurate visual
  explanation of {CNN}s for {SAR} image interpretation}.
\newblock \bibinfo{journal}{Remote Sensing} \bibinfo{volume}{13},
  \bibinfo{pages}{1772}.
%Type = Inproceedings
\bibitem[{Fu et~al.(2020)Fu, Hu, Dong, Guo, Gao and Li}]{Xgradcam}
\bibinfo{author}{Fu, R.}, \bibinfo{author}{Hu, Q.}, \bibinfo{author}{Dong, X.},
  \bibinfo{author}{Guo, Y.}, \bibinfo{author}{Gao, Y.}, \bibinfo{author}{Li,
  B.}, \bibinfo{year}{2020}.
\newblock \bibinfo{title}{Axiom-based {G}rad-{CAM}: Towards accurate
  visualization and explanation of {CNN}s}, in: \bibinfo{booktitle}{In
  Proceedings of the 2020 British Machine Vision Conference (BMVC 2020)}.
%Type = Inproceedings
\bibitem[{He et~al.(2016)He, Zhang, Ren and Sun}]{ResNet}
\bibinfo{author}{He, K.}, \bibinfo{author}{Zhang, X.}, \bibinfo{author}{Ren,
  S.}, \bibinfo{author}{Sun, J.}, \bibinfo{year}{2016}.
\newblock \bibinfo{title}{Deep residual learning for image recognition}, in:
  \bibinfo{booktitle}{In Proceedings of 2016 IEEE conference on Computer Vision
  and Pattern Recognition (CVPR)}, pp. \bibinfo{pages}{770--778}.
\newblock \DOIprefix\doi{10.1109/CVPR.2016.90}.
%Type = Inproceedings
\bibitem[{Krizhevsky et~al.(2012)Krizhevsky, Sutskever and Hinton}]{AlexNet}
\bibinfo{author}{Krizhevsky, A.}, \bibinfo{author}{Sutskever, I.},
  \bibinfo{author}{Hinton, G.E.}, \bibinfo{year}{2012}.
\newblock \bibinfo{title}{Imagenet classification with deep convolutional
  neural networks}, in: \bibinfo{editor}{Pereira, F.}, \bibinfo{editor}{Burges,
  C.}, \bibinfo{editor}{Bottou, L.}, \bibinfo{editor}{Weinberger, K.} (Eds.),
  \bibinfo{booktitle}{Advances in Neural Information Processing Systems},
  \bibinfo{publisher}{Curran Associates, Inc.}
%Type = Article
\bibitem[{Lapuschkin et~al.(2019)Lapuschkin, W{\"a}ldchen, Binder, Montavon,
  Samek and M{\"u}ller}]{CleverHans}
\bibinfo{author}{Lapuschkin, S.}, \bibinfo{author}{W{\"a}ldchen, S.},
  \bibinfo{author}{Binder, A.}, \bibinfo{author}{Montavon, G.},
  \bibinfo{author}{Samek, W.}, \bibinfo{author}{M{\"u}ller, K.R.},
  \bibinfo{year}{2019}.
\newblock \bibinfo{title}{Unmasking clever hans predictors and assessing what
  machines really learn}.
\newblock \bibinfo{journal}{Nature communications} \bibinfo{volume}{10},
  \bibinfo{pages}{1--8}.
%Type = Article
\bibitem[{Liang et~al.(2018)Liang, Hu, Zhang, Lin and Xing}]{liang2018symbolic}
\bibinfo{author}{Liang, X.}, \bibinfo{author}{Hu, Z.}, \bibinfo{author}{Zhang,
  H.}, \bibinfo{author}{Lin, L.}, \bibinfo{author}{Xing, E.P.},
  \bibinfo{year}{2018}.
\newblock \bibinfo{title}{Symbolic graph reasoning meets convolutions}.
\newblock \bibinfo{journal}{Advances in Neural Information Processing Systems}
  \bibinfo{volume}{31}.
%Type = Article
\bibitem[{Liu et~al.(2023)Liu, Zhang, Zhou and Wang}]{LIU202327}
\bibinfo{author}{Liu, J.}, \bibinfo{author}{Zhang, F.}, \bibinfo{author}{Zhou,
  Z.}, \bibinfo{author}{Wang, J.}, \bibinfo{year}{2023}.
\newblock \bibinfo{title}{Bfmnet: Bilateral feature fusion network with
  multi-scale context aggregation for real-time semantic segmentation}.
\newblock \bibinfo{journal}{Neurocomputing} \bibinfo{volume}{521},
  \bibinfo{pages}{27--40}.
\newblock \DOIprefix\doi{https://doi.org/10.1016/j.neucom.2022.11.084}.
%Type = Article
\bibitem[{Liu et~al.(2022a)Liu, Meng, Li, Mao and Chen}]{LIU2022193}
\bibinfo{author}{Liu, K.}, \bibinfo{author}{Meng, R.}, \bibinfo{author}{Li,
  L.}, \bibinfo{author}{Mao, J.}, \bibinfo{author}{Chen, H.},
  \bibinfo{year}{2022}a.
\newblock \bibinfo{title}{Sisl-net: Saliency-guided self-supervised learning
  network for image classification}.
\newblock \bibinfo{journal}{Neurocomputing} \bibinfo{volume}{510},
  \bibinfo{pages}{193--202}.
\newblock \DOIprefix\doi{https://doi.org/10.1016/j.neucom.2022.09.029}.
%Type = Inproceedings
\bibitem[{Liu et~al.(2022b)Liu, Mao, Wu, Feichtenhofer, Darrell and
  Xie}]{liu2022convnet}
\bibinfo{author}{Liu, Z.}, \bibinfo{author}{Mao, H.}, \bibinfo{author}{Wu,
  C.Y.}, \bibinfo{author}{Feichtenhofer, C.}, \bibinfo{author}{Darrell, T.},
  \bibinfo{author}{Xie, S.}, \bibinfo{year}{2022}b.
\newblock \bibinfo{title}{A convnet for the 2020s}, in: \bibinfo{booktitle}{In
  Proceedings of the IEEE/CVF Conference on Computer Vision and Pattern
  Recognition (CVPR)}, pp. \bibinfo{pages}{11976--11986}.
%Type = Article
\bibitem[{Ma et~al.(2021)Ma, Zhang, Pena-Pena and Arce}]{MA2021108301}
\bibinfo{author}{Ma, X.}, \bibinfo{author}{Zhang, S.},
  \bibinfo{author}{Pena-Pena, K.}, \bibinfo{author}{Arce, G.R.},
  \bibinfo{year}{2021}.
\newblock \bibinfo{title}{Fast spectral clustering method based on graph
  similarity matrix completion}.
\newblock \bibinfo{journal}{Signal Processing} \bibinfo{volume}{189},
  \bibinfo{pages}{108301}.
\newblock \DOIprefix\doi{https://doi.org/10.1016/j.sigpro.2021.108301}.
%Type = Article
\bibitem[{Macpherson et~al.(2021)Macpherson, Churchland, Sejnowski, DiCarlo,
  Kamitani, Takahashi and Hikida}]{MACPHERSON2021603}
\bibinfo{author}{Macpherson, T.}, \bibinfo{author}{Churchland, A.},
  \bibinfo{author}{Sejnowski, T.}, \bibinfo{author}{DiCarlo, J.},
  \bibinfo{author}{Kamitani, Y.}, \bibinfo{author}{Takahashi, H.},
  \bibinfo{author}{Hikida, T.}, \bibinfo{year}{2021}.
\newblock \bibinfo{title}{Natural and artificial intelligence: A brief
  introduction to the interplay between ai and neuroscience research}.
\newblock \bibinfo{journal}{Neural Networks} \bibinfo{volume}{144},
  \bibinfo{pages}{603--613}.
\newblock \DOIprefix\doi{https://doi.org/10.1016/j.neunet.2021.09.018}.
%Type = Article
\bibitem[{Omeiza et~al.(2019)Omeiza, Speakman, Cintas and
  Weldermariam}]{omeiza2019smooth}
\bibinfo{author}{Omeiza, D.}, \bibinfo{author}{Speakman, S.},
  \bibinfo{author}{Cintas, C.}, \bibinfo{author}{Weldermariam, K.},
  \bibinfo{year}{2019}.
\newblock \bibinfo{title}{Smooth grad-cam++: An enhanced inference level
  visualization technique for deep convolutional neural network models}.
\newblock \bibinfo{journal}{arXiv preprint arXiv:1908.01224} .
%Type = Inproceedings
\bibitem[{Ramaswamy et~al.(2020)}]{AblationCAM}
\bibinfo{author}{Ramaswamy, H.G.}, et~al., \bibinfo{year}{2020}.
\newblock \bibinfo{title}{Ablation-cam: Visual explanations for deep
  convolutional network via gradient-free localization}, in:
  \bibinfo{booktitle}{In Proceedings of the IEEE Winter Conference on
  Applications of Computer Vision (WACV)}, pp. \bibinfo{pages}{983--991}.
%Type = Inproceedings
\bibitem[{Redmon et~al.(2016)Redmon, Divvala, Girshick and Farhadi}]{Yolo}
\bibinfo{author}{Redmon, J.}, \bibinfo{author}{Divvala, S.},
  \bibinfo{author}{Girshick, R.}, \bibinfo{author}{Farhadi, A.},
  \bibinfo{year}{2016}.
\newblock \bibinfo{title}{You only look once: Unified, real-time object
  detection}, in: \bibinfo{booktitle}{In Proceedings of 2016 IEEE Conference on
  Computer Vision and Pattern Recognition (CVPR)}, pp.
  \bibinfo{pages}{779--788}.
\newblock \DOIprefix\doi{10.1109/CVPR.2016.91}.
%Type = Inproceedings
\bibitem[{Ren et~al.(2021)Ren, Li, Liu and Zhang}]{ren2021interpreting}
\bibinfo{author}{Ren, J.}, \bibinfo{author}{Li, M.}, \bibinfo{author}{Liu, Z.},
  \bibinfo{author}{Zhang, Q.}, \bibinfo{year}{2021}.
\newblock \bibinfo{title}{Interpreting and disentangling feature components of
  various complexity from {DNN}s}, in: \bibinfo{booktitle}{In proceedings of
  International Conference on Machine Learning}, \bibinfo{organization}{PMLR}.
  pp. \bibinfo{pages}{8971--8981}.
%Type = Article
\bibitem[{Saleem et~al.(2022)Saleem, Yuan, Kurugollu, Anjum and
  Liu}]{SALEEM2022165}
\bibinfo{author}{Saleem, R.}, \bibinfo{author}{Yuan, B.},
  \bibinfo{author}{Kurugollu, F.}, \bibinfo{author}{Anjum, A.},
  \bibinfo{author}{Liu, L.}, \bibinfo{year}{2022}.
\newblock \bibinfo{title}{Explaining deep neural networks: A survey on the
  global interpretation methods}.
\newblock \bibinfo{journal}{Neurocomputing} \bibinfo{volume}{513},
  \bibinfo{pages}{165--180}.
\newblock \DOIprefix\doi{https://doi.org/10.1016/j.neucom.2022.09.129}.
%Type = Article
\bibitem[{Scalzo et~al.(2023)Scalzo, Stanković, Daković, Constantinides and
  Mandic}]{SCALZO202383}
\bibinfo{author}{Scalzo, B.}, \bibinfo{author}{Stanković, L.},
  \bibinfo{author}{Daković, M.}, \bibinfo{author}{Constantinides, A.G.},
  \bibinfo{author}{Mandic, D.P.}, \bibinfo{year}{2023}.
\newblock \bibinfo{title}{A class of doubly stochastic shift operators for
  random graph signals and their boundedness}.
\newblock \bibinfo{journal}{Neural Networks} \bibinfo{volume}{158},
  \bibinfo{pages}{83--88}.
\newblock \DOIprefix\doi{https://doi.org/10.1016/j.neunet.2022.10.035}.
%Type = Inproceedings
\bibitem[{Selvaraju et~al.(2017)Selvaraju, Cogswell, Das, Vedantam, Parikh and
  Batra}]{GradCAM}
\bibinfo{author}{Selvaraju, R.R.}, \bibinfo{author}{Cogswell, M.},
  \bibinfo{author}{Das, A.}, \bibinfo{author}{Vedantam, R.},
  \bibinfo{author}{Parikh, D.}, \bibinfo{author}{Batra, D.},
  \bibinfo{year}{2017}.
\newblock \bibinfo{title}{Grad-{CAM}: Visual explanations from deep networks
  via gradient-based localization}, in: \bibinfo{booktitle}{In Proceedings of
  the 2017 IEEE international conference on computer vision}, pp.
  \bibinfo{pages}{618--626}.
%Type = Article
\bibitem[{Simonyan et~al.(2013)Simonyan, Vedaldi and
  Zisserman}]{simonyan2013deep}
\bibinfo{author}{Simonyan, K.}, \bibinfo{author}{Vedaldi, A.},
  \bibinfo{author}{Zisserman, A.}, \bibinfo{year}{2013}.
\newblock \bibinfo{title}{Deep inside convolutional networks: Visualising image
  classification models and saliency maps}.
\newblock \bibinfo{journal}{arXiv preprint arXiv:1312.6034} .
%Type = Inproceedings
\bibitem[{Simonyan and Zisserman.(2015)}]{VGG16}
\bibinfo{author}{Simonyan, K.}, \bibinfo{author}{Zisserman., A.},
  \bibinfo{year}{2015}.
\newblock \bibinfo{title}{Very deep convolutional networks for large-scale
  image recognition}, in: \bibinfo{booktitle}{3rd International Conference on
  Learning Representations (ICLR 2015)}, pp. \bibinfo{pages}{1--14}.
%Type = Article
\bibitem[{Spinelli et~al.(2022)Spinelli, Scardapane and Uncini}]{9772740}
\bibinfo{author}{Spinelli, I.}, \bibinfo{author}{Scardapane, S.},
  \bibinfo{author}{Uncini, A.}, \bibinfo{year}{2022}.
\newblock \bibinfo{title}{A meta-learning approach for training explainable
  graph neural networks}.
\newblock \bibinfo{journal}{IEEE Transactions on Neural Networks and Learning
  Systems} , \bibinfo{pages}{1--9}\DOIprefix\doi{10.1109/TNNLS.2022.3171398}.
%Type = Inproceedings
\bibitem[{Srinivas et~al.(2021)Srinivas, Lin, Parmar, Shlens, Abbeel and
  Vaswani}]{srinivas2021bottleneck}
\bibinfo{author}{Srinivas, A.}, \bibinfo{author}{Lin, T.Y.},
  \bibinfo{author}{Parmar, N.}, \bibinfo{author}{Shlens, J.},
  \bibinfo{author}{Abbeel, P.}, \bibinfo{author}{Vaswani, A.},
  \bibinfo{year}{2021}.
\newblock \bibinfo{title}{Bottleneck transformers for visual recognition}, in:
  \bibinfo{booktitle}{In Proceedings of the IEEE/CVF Conference on Computer
  Vision and Pattern recognition (CVPR)}, pp. \bibinfo{pages}{16519--16529}.
%Type = Article
\bibitem[{Stankovic et~al.(2017)Stankovic, Dakovic and Sejdic}]{spm}
\bibinfo{author}{Stankovic, L.}, \bibinfo{author}{Dakovic, M.},
  \bibinfo{author}{Sejdic, E.}, \bibinfo{year}{2017}.
\newblock \bibinfo{title}{Vertex-frequency analysis: A way to localize graph
  spectral components [lecture notes]}.
\newblock \bibinfo{journal}{IEEE Signal Processing Magazine}
  \bibinfo{volume}{34}, \bibinfo{pages}{176--182}.
\newblock \DOIprefix\doi{10.1109/MSP.2017.2696572}.
%Type = Article
\bibitem[{Stankovic et~al.(2019)Stankovic, Mandic, Dakovic, Kisil, Sejdic and
  Constantinides}]{SPM2}
\bibinfo{author}{Stankovic, L.}, \bibinfo{author}{Mandic, D.P.},
  \bibinfo{author}{Dakovic, M.}, \bibinfo{author}{Kisil, I.},
  \bibinfo{author}{Sejdic, E.}, \bibinfo{author}{Constantinides, A.G.},
  \bibinfo{year}{2019}.
\newblock \bibinfo{title}{Understanding the basis of graph signal processing
  via an intuitive example-driven approach}.
\newblock \bibinfo{journal}{IEEE Signal Processing Magazine}
  \bibinfo{volume}{36}, \bibinfo{pages}{133--145}.
\newblock \DOIprefix\doi{10.1109/MSP.2019.2929832}.
%Type = Article
\bibitem[{Sun et~al.(2022)Sun, Song, Cai, Du and Guizani}]{SUN2022989}
\bibinfo{author}{Sun, S.}, \bibinfo{author}{Song, B.}, \bibinfo{author}{Cai,
  X.}, \bibinfo{author}{Du, X.}, \bibinfo{author}{Guizani, M.},
  \bibinfo{year}{2022}.
\newblock \bibinfo{title}{{CAMA}: Class activation mapping disruptive attack
  for deep neural networks}.
\newblock \bibinfo{journal}{Neurocomputing} \bibinfo{volume}{500},
  \bibinfo{pages}{989--1002}.
\newblock \DOIprefix\doi{https://doi.org/10.1016/j.neucom.2022.05.065}.
%Type = Article
\bibitem[{Tan et~al.(2022)Tan, Gao, Khan and Guan}]{TAN202258}
\bibinfo{author}{Tan, R.}, \bibinfo{author}{Gao, L.}, \bibinfo{author}{Khan,
  N.}, \bibinfo{author}{Guan, L.}, \bibinfo{year}{2022}.
\newblock \bibinfo{title}{Interpretable artificial intelligence through
  locality guided neural networks}.
\newblock \bibinfo{journal}{Neural Networks} \bibinfo{volume}{155},
  \bibinfo{pages}{58--73}.
\newblock \DOIprefix\doi{https://doi.org/10.1016/j.neunet.2022.08.009}.
%Type = Article
\bibitem[{Townsend et~al.(2020)Townsend, Chaton and Monteiro}]{8889997}
\bibinfo{author}{Townsend, J.}, \bibinfo{author}{Chaton, T.},
  \bibinfo{author}{Monteiro, J.M.}, \bibinfo{year}{2020}.
\newblock \bibinfo{title}{Extracting relational explanations from deep neural
  networks: A survey from a neural-symbolic perspective}.
\newblock \bibinfo{journal}{IEEE Transactions on Neural Networks and Learning
  Systems} \bibinfo{volume}{31}, \bibinfo{pages}{3456--3470}.
\newblock \DOIprefix\doi{10.1109/TNNLS.2019.2944672}.
%Type = Article
\bibitem[{Tu et~al.(2021)Tu, Zhou, Gan, Jiang, Hussain and Luo}]{TU2021443}
\bibinfo{author}{Tu, Z.}, \bibinfo{author}{Zhou, A.}, \bibinfo{author}{Gan,
  C.}, \bibinfo{author}{Jiang, B.}, \bibinfo{author}{Hussain, A.},
  \bibinfo{author}{Luo, B.}, \bibinfo{year}{2021}.
\newblock \bibinfo{title}{A novel domain activation mapping-guided network
  ({DA-GNT}) for visual tracking}.
\newblock \bibinfo{journal}{Neurocomputing} \bibinfo{volume}{449},
  \bibinfo{pages}{443--454}.
\newblock \DOIprefix\doi{https://doi.org/10.1016/j.neucom.2021.03.056}.
%Type = Article
\bibitem[{Vlahek and Mongus(2021)}]{9528915}
\bibinfo{author}{Vlahek, D.}, \bibinfo{author}{Mongus, D.},
  \bibinfo{year}{2021}.
\newblock \bibinfo{title}{An efficient iterative approach to explainable
  feature learning}.
\newblock \bibinfo{journal}{IEEE Transactions on Neural Networks and Learning
  Systems} , \bibinfo{pages}{1--13}\DOIprefix\doi{10.1109/TNNLS.2021.3107049}.
%Type = Inproceedings
\bibitem[{Wang et~al.(2020)Wang, Wang, Du, Yang, Zhang, Ding, Mardziel and
  Hu}]{ScoreCAM}
\bibinfo{author}{Wang, H.}, \bibinfo{author}{Wang, Z.}, \bibinfo{author}{Du,
  M.}, \bibinfo{author}{Yang, F.}, \bibinfo{author}{Zhang, Z.},
  \bibinfo{author}{Ding, S.}, \bibinfo{author}{Mardziel, P.},
  \bibinfo{author}{Hu, X.}, \bibinfo{year}{2020}.
\newblock \bibinfo{title}{Score-{CAM}: Score-weighted visual explanations for
  convolutional neural networks}, in: \bibinfo{booktitle}{In Proceedings of the
  IEEE Conference on Computer Vision and Pattern Recognition (CVPR) workshops},
  pp. \bibinfo{pages}{24--25}.
%Type = Inproceedings
\bibitem[{Zeiler and Fergus(2014)}]{zeiler2014visualizing}
\bibinfo{author}{Zeiler, M.D.}, \bibinfo{author}{Fergus, R.},
  \bibinfo{year}{2014}.
\newblock \bibinfo{title}{Visualizing and understanding convolutional
  networks}, in: \bibinfo{booktitle}{European conference on computer vision},
  \bibinfo{organization}{Springer}. pp. \bibinfo{pages}{818--833}.
%Type = Article
\bibitem[{Zhang et~al.(2021)Zhang, Rao and Yang}]{zhang2021group}
\bibinfo{author}{Zhang, Q.}, \bibinfo{author}{Rao, L.}, \bibinfo{author}{Yang,
  Y.}, \bibinfo{year}{2021}.
\newblock \bibinfo{title}{Group-{CAM}: {G}roup score-weighted visual
  explanations for deep convolutional networks}.
\newblock \bibinfo{journal}{arXiv preprint arXiv:2103.13859} .
%Type = Article
\bibitem[{Zhao et~al.(2019)Zhao, Xie, Wang, Liu, Shi and Du}]{ZHAO2019185}
\bibinfo{author}{Zhao, Z.}, \bibinfo{author}{Xie, X.}, \bibinfo{author}{Wang,
  C.}, \bibinfo{author}{Liu, W.}, \bibinfo{author}{Shi, G.},
  \bibinfo{author}{Du, J.}, \bibinfo{year}{2019}.
\newblock \bibinfo{title}{Visualizing and understanding of learned compressive
  sensing with residual network}.
\newblock \bibinfo{journal}{Neurocomputing} \bibinfo{volume}{359},
  \bibinfo{pages}{185--198}.
\newblock \DOIprefix\doi{https://doi.org/10.1016/j.neucom.2019.05.043}.
%Type = Article
\bibitem[{Zhou et~al.(2019)Zhou, Bau, Oliva and Torralba}]{NETWORKDISSECTION}
\bibinfo{author}{Zhou, B.}, \bibinfo{author}{Bau, D.}, \bibinfo{author}{Oliva,
  A.}, \bibinfo{author}{Torralba, A.}, \bibinfo{year}{2019}.
\newblock \bibinfo{title}{Interpreting deep visual representations via network
  dissection}.
\newblock \bibinfo{journal}{IEEE Transactions on Pattern Analysis and Machine
  Intelligence} \bibinfo{volume}{41}, \bibinfo{pages}{2131--2145}.
\newblock \DOIprefix\doi{10.1109/TPAMI.2018.2858759}.
%Type = Inproceedings
\bibitem[{Zhou et~al.(2016)Zhou, Khosla, Lapedriza, Oliva and Torralba}]{CAM}
\bibinfo{author}{Zhou, B.}, \bibinfo{author}{Khosla, A.},
  \bibinfo{author}{Lapedriza, A.}, \bibinfo{author}{Oliva, A.},
  \bibinfo{author}{Torralba, A.}, \bibinfo{year}{2016}.
\newblock \bibinfo{title}{Learning deep features for discriminative
  localization}, in: \bibinfo{booktitle}{In Proceedings of the 2016 IEEE
  Conference on Computer Vision and Pattern Recognition (CVPR)}, pp.
  \bibinfo{pages}{2921--2929}.
%Type = Article
\bibitem[{Zhou and Kainz(2018)}]{zhou2018efficient}
\bibinfo{author}{Zhou, K.}, \bibinfo{author}{Kainz, B.}, \bibinfo{year}{2018}.
\newblock \bibinfo{title}{Efficient image evidence analysis of cnn
  classification results}.
\newblock \bibinfo{journal}{arXiv preprint arXiv:1801.01693} .
%Type = Article
\bibitem[{Zhu et~al.(2022)Zhu, Feng, Stankovi{\'c}, Ding, Fan and
  Zhou}]{ProbeFeature}
\bibinfo{author}{Zhu, M.}, \bibinfo{author}{Feng, Z.},
  \bibinfo{author}{Stankovi{\'c}, L.}, \bibinfo{author}{Ding, L.},
  \bibinfo{author}{Fan, J.}, \bibinfo{author}{Zhou, X.}, \bibinfo{year}{2022}.
\newblock \bibinfo{title}{A probe-feature for specific emitter identification
  using axiom-based grad-cam}.
\newblock \bibinfo{journal}{Signal Processing} , \bibinfo{pages}{108685}.
%Type = Inproceedings
\bibitem[{Zhu et~al.(2017)Zhu, Zhao, Wang, Zhao, Wu and Lu}]{zhu2017couplenet}
\bibinfo{author}{Zhu, Y.}, \bibinfo{author}{Zhao, C.}, \bibinfo{author}{Wang,
  J.}, \bibinfo{author}{Zhao, X.}, \bibinfo{author}{Wu, Y.},
  \bibinfo{author}{Lu, H.}, \bibinfo{year}{2017}.
\newblock \bibinfo{title}{Couplenet: Coupling global structure with local parts
  for object detection}, in: \bibinfo{booktitle}{Proceedings of the IEEE
  international conference on computer vision}, pp.
  \bibinfo{pages}{4126--4134}.

\end{thebibliography}


%Author biography without author photo.
\bio{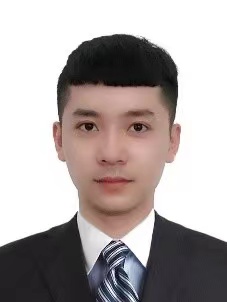}
\textbf{Zhenpeng Feng} 
was born in Xianyang, Shaanxi, China in 1996. He received a B.E. degree in School of Electronic Engineering, Xidian University in 2019. He is currently a Ph.D. student in explainable artificial intelligence at School of Electronic Engineering, Xidian University. He is also a visiting student in the University of Montenegro, working with Prof. Ljubi\v{s}a Stankovi\'c's research team.  His research interests include interpreting deep neural networks and signal processing. 
\endbio

\bio{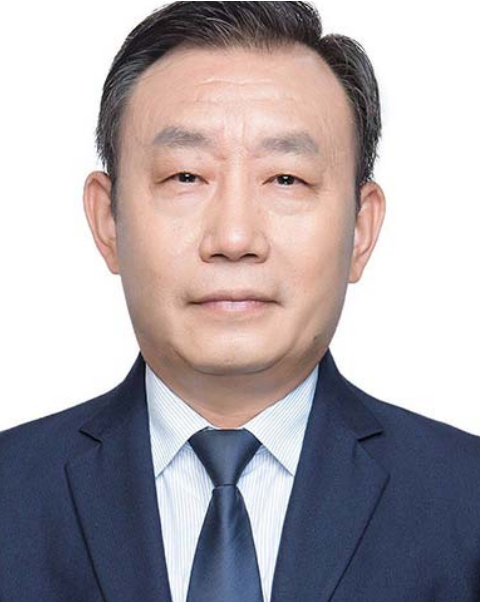}
\textbf{Hongbing Ji} 
received a B.S. degree in radar engineering, an M.S. degree in
circuit, signals, and systems, and the Ph.D. degree in signal and information processing from Xidian University, Xi’an, China, in 1983, 1989, and 1999, respectively. He is currently a full professor at Xidian University and a senior member of IEEE. His research interests include pattern recognition, radar signal processing, and multi-sensor information fusion. 
\endbio

\newpage

\bio{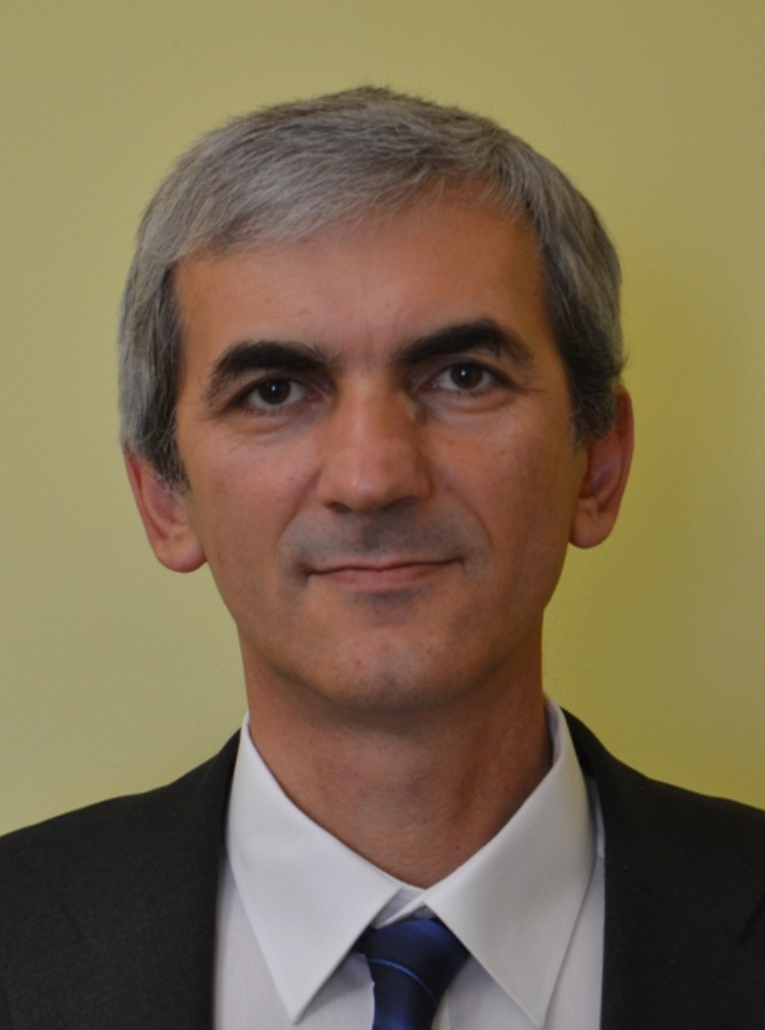}
\textbf{Milo\v s Dakovi\' c} 
was born in 1970 in Nik\v si\' c, Montenegro. He received a B.S. in 1996, an M.S. in 2001, and a Ph.D. in 2005, all in electrical engineering from the University of Montenegro. He is a full professor at the University of Montenegro. His research interests are in signal processing, time-frequency signal analysis, compressive sensing, radar signal processing, and graph signal processing.
\endbio

\bio{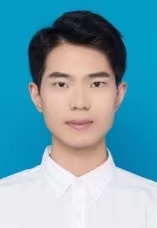}
\textbf{Xiyang Cui} 
was born in Handan, Hebei, China in 1997. He received the B.E. degree and M.E. degree in Electronic Information Engineering and Electrical Circuit System from School of Electronic Engineering, Xidian University in 2019 and 2021, respectively. He is currently an investigator of an electronic company and collaborates with Zhenpeng Feng and Prof. Ljubi\v sa Stankovi\'c in scientific research.  His research interests include electrical circuit design and image processing.
\endbio

\bio{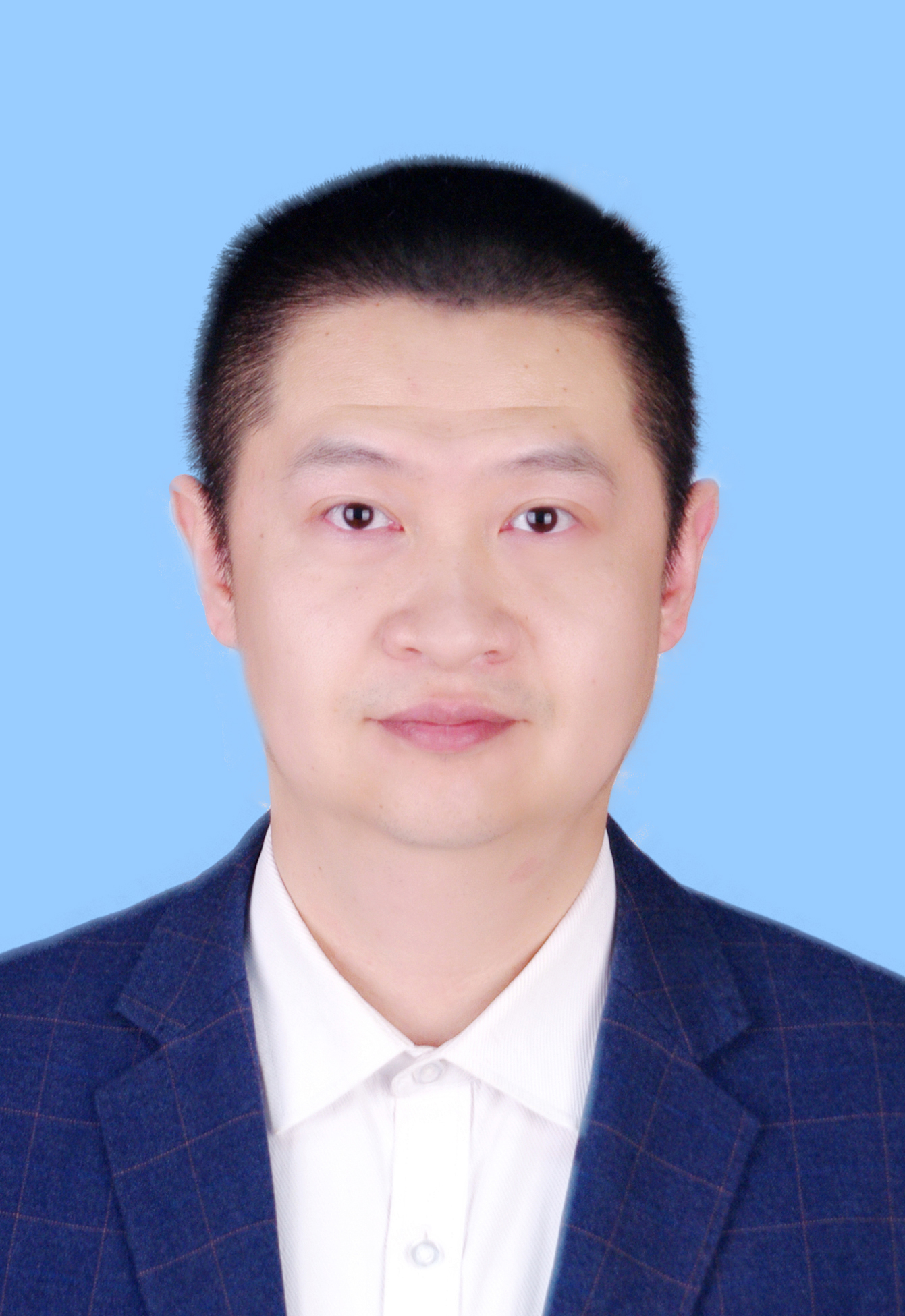}
\textbf{Mingzhe Zhu} 
was born in China in 1982.
He received a B.S. degree in signal and information processing, a Ph.D. degree in pattern recognition and intelligent system from Xidian University in 2004 and 2010, respectively. He is currently an associate professor at School of Electronic Engineering, Xidian University. His research interests include non-stationary signal processing, time-frequency analysis, and target recognition.
\endbio
\medskip

\bio{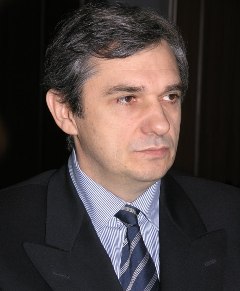}
\textbf{Ljubi\v{s}a Stankovi\'{c}}  was born in Montenegro, 1960. He was at the Ruhr University Bochum, 1997-1999, supported by the AvH Foundation. Stankovic was the Rector of the University of Montenegro 2003-2008, the Ambassador of Montenegro to the UK, 2011-2015, and a visiting academic to the Imperial College London, 2012-2013.  He published almost 200 journal papers. He is a member of the National Academy of Science and Arts (CANU) and the Academia Europaea. Stankovi\'c won the Best paper award from the EURASIP in 2017 and the IEEE SPM Best Column Award for 2020. Stankovi\'c is a professor at the University of Montenegro and a Fellow of the IEEE.
\endbio

\end{document}